\pdfoutput=1
\documentclass[11pt]{article}
\usepackage{style/depth}
\usepackage{graphicx}
\usepackage{xspace}
\usepackage{placeins}

\newcommand{\Dstar}{\ensuremath{D^{*}}}
\newcommand{\Wstar}{\ensuremath{W^{*}}}
\newcommand{\ours}{\texorpdfstring{\textsc{Depth}}{DEPTH}\xspace}

\newenvironment{widefigure}[1][t]{\begin{figure}[#1]}{\end{figure}}
\newenvironment{widetable}[1][t]{\begin{table}[#1]}{\end{table}}

\title{Is Your Long-Horizon Benchmark Really Deep?\\ Measuring and Synthesizing Depth in Terminal Tasks}
\author{Yu~Jiang\textsuperscript{1,*}\quad Yue~Huang\textsuperscript{1,*}\quad Hang~Hua\textsuperscript{2,3}\quad
  Yuyang~Wu\textsuperscript{4}\quad Zixuan~Weng\textsuperscript{5}\quad Olexandr~Isayev\textsuperscript{4}\quad
  Yue~Zhao\textsuperscript{6}\quad Zinan~Lin\textsuperscript{7}\quad Xiangliang~Zhang\textsuperscript{1}}
\definecolor{cmured}{HTML}{C41230}
\newcommand{\affillogo}[1]{\raisebox{\dimexpr0.34em-0.5\height\relax}{%
  \includegraphics[width=2em,height=1em,keepaspectratio]{figures/affiliations/#1.pdf}}}
\newcommand{\affilitem}[3]{\makebox[2.3em][c]{#1}\hspace{0.35em}\textsuperscript{#2}#3}
\newcommand{\titleaffiliations}{{\Affilfont\renewcommand{\arraystretch}{1.5}%
  \begin{tabular}{@{}l@{\hspace{3.5em}}l@{}}
    \affilitem{\affillogo{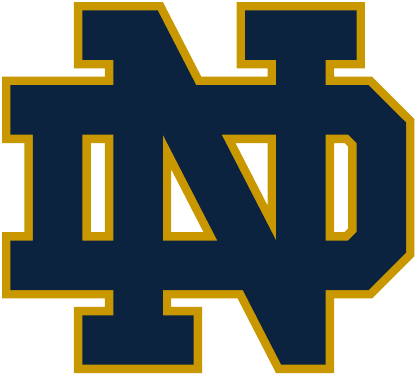}}{1}{University of Notre Dame} &
    \affilitem{\affillogo{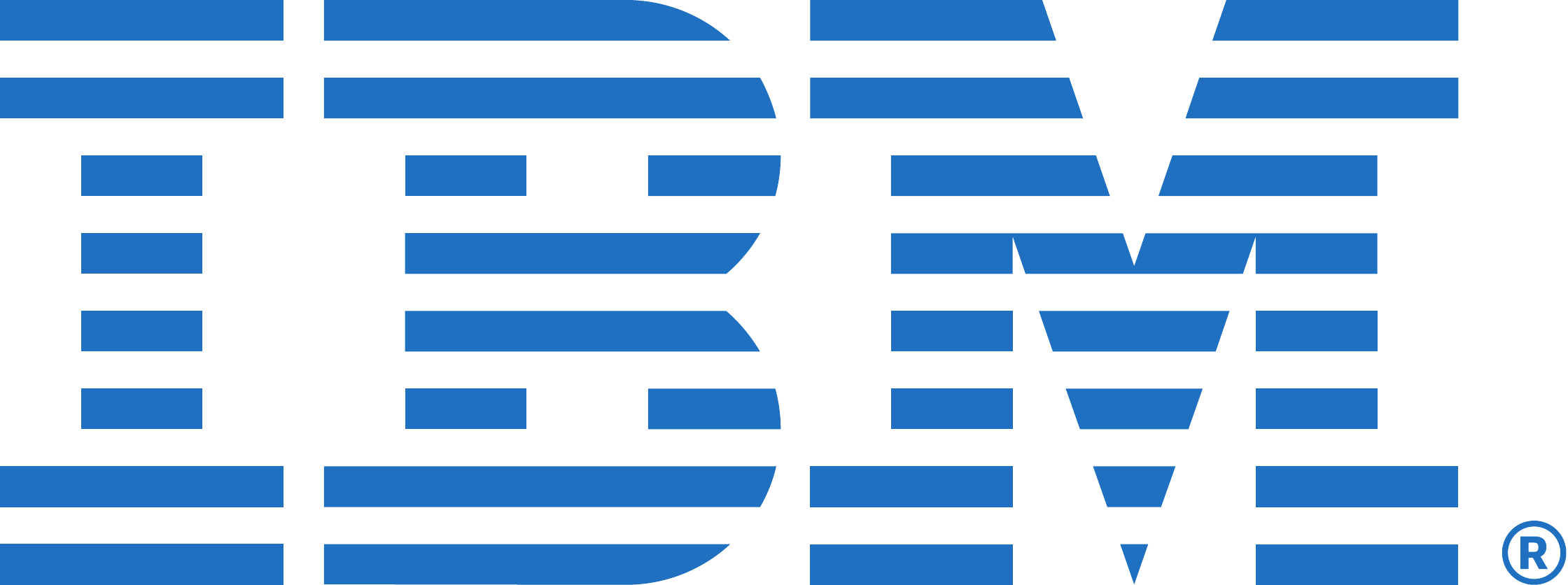}}{2}{IBM Research} \\
    \affilitem{\affillogo{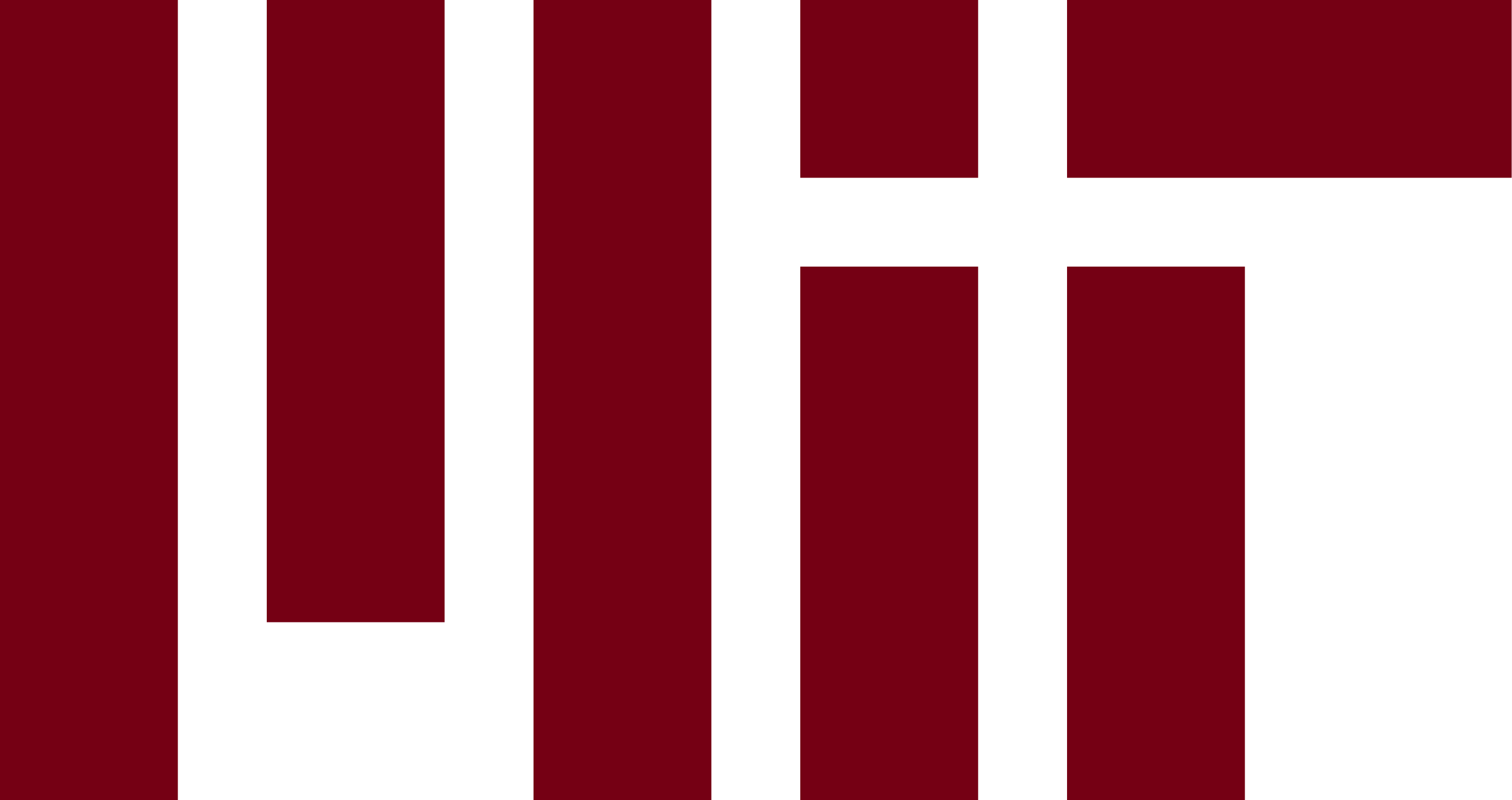}}{3}{MIT-IBM Watson AI Lab} &
    \affilitem{{\color{cmured}\fontsize{8.3}{8.3}\selectfont\bfseries CMU}}{4}{Carnegie Mellon University} \\
    \affilitem{\affillogo{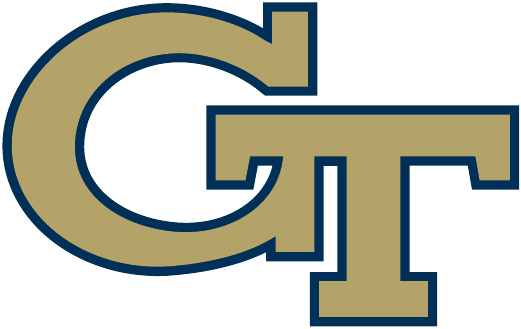}}{5}{Georgia Institute of Technology} &
    \affilitem{\affillogo{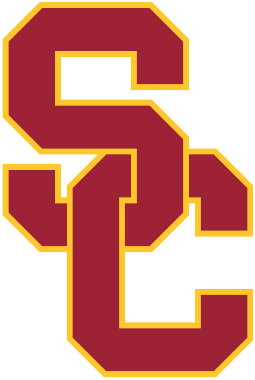}}{6}{University of Southern California} \\
    \affilitem{\affillogo{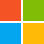}}{7}{Microsoft Research}
  \end{tabular}\par\vskip2pt\textsuperscript{*}Equal contribution}}
\date{}

\hypersetup{pdftitle={Is Your Long-Horizon Benchmark Really Deep? Measuring and Synthesizing Depth in Terminal Tasks}}

\begin{document}
\maketitle
\thispagestyle{titlepage}

\begin{abstract}
Long-horizon terminal tasks are used to evaluate and train agents. Human-curated datasets are scarce, so many synthesis methods have emerged. They usually measure the length of a task by the turns a model needs to pass it, and its difficulty by labels the LLM sets during generation. We find, however, that the turn count largely reflects the habits of the model, not only the demands of the task, and that the labels barely track how often models pass. We describe a terminal task by its work, the number of operations it requires, and its depth, the number of turns that must happen in sequence because the commands of each turn can only be chosen after reading the output of the previous one. Estimating depth as the smallest turn cap, stated to the agent, under which the task passes, we audit 734 tasks from six existing datasets. Synthesis has mostly added work, not depth: 94\% of the solved synthetic tasks pass within three turns, yet depth, unlike the turn count, predicts difficulty beyond work. We propose \ours, in which a generation agent writes each task one segment at a time in a real software environment, carrying out each segment before writing it and starting the next from the state it leaves. Our 105 tasks exceed every audited dataset in depth and work, and none of six current models passes half of them. Fine-tuning Qwen3-8B on runs from these tasks improves it on two terminal benchmarks.
\end{abstract}

{\centering\vspace*{-12pt}
  \href{https://github.com/AlanRosston/Depth}{\raisebox{-0.17em}{\includegraphics[height=1.1em]{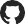}}%
  \hspace{0.4em}Code}\hspace{2.5em}%
  \href{https://huggingface.co/datasets/AlanRosston520/Depth}{\raisebox{-0.17em}{\includegraphics[height=1.1em]{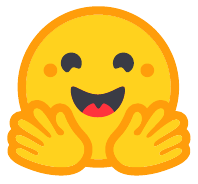}}%
  \hspace{0.4em}Dataset}\par\vspace{14pt}}
\section{Introduction}
\label{sec:intro}

An agent working in a terminal issues shell commands in each turn, reads their output and decides what to run next, which lets it install software, configure services and debug programs~\citep{merrill2026terminalbench}. Completing long tasks reliably, however, remains a challenge~\citep{kwa2025measuring,sinha2026illusion}: the agent must keep track of what it has learned about the environment over many turns, and an early mistake can derail every later step.

Long-horizon terminal tasks are therefore needed both to evaluate agents and to train them. Each task bundles an instruction, an environment and an automatic verifier~\citep{merrill2026terminalbench,harbor2026}, and training needs such tasks in large numbers~\citep{pi2026nemotron,raoof2026openthoughts,peng2026litecoderterminal}. Two kinds of methods provide them. Human-curated sets are written or reviewed by people: Terminal-Bench~2.0 (TB2;~\citealp{merrill2026terminalbench}) consists of tasks written by people, and TerminalWorld-Verified (TW;~\citealp{chu2026terminalworld}) builds tasks from real terminal recordings and has people review them. Synthesis methods generate tasks automatically~\citep{cheng2026terminalworld}: TMax~\citep{ivison2026tmax} and SETA~\citep{shen2026seta} generate tasks in bulk with controlled difficulty, CLI-Gym~\citep{lin2026cligym} derives repair tasks by breaking Dockerfiles, FACET~\citep{shi2026facet} composes skill packages into scenarios, and RST~\citep{li2026rst} extends verified tasks over recursive rounds, each perturbing the environment and rewriting the instruction and verifier.

Both kinds have limitations. Human-curated sets are reliable but small and costly to extend, far from the numbers that training needs. Synthesis methods produce tasks in any number, but it is unclear whether these tasks are long in the sense that matters. They, like recent long-horizon benchmarks~\citep{li2026longhorizonterminalbench}, report length as the number of turns or steps a model takes under a loose turn limit~\citep{ivison2026tmax,shi2026facet,li2026rst,zou2026scaling}, a count that reflects the habits of the model as much as the demands of the task~\citep{meng2026calibforge,desai2026swemarathon}. Difficulty is given by labels set during generation~\citep{ivison2026tmax,shen2026seta}, which record what was requested rather than what models find hard. Most of these pipelines verify only that each new task is solvable. Pipelines that aim for later steps to build on earlier ones obtain this by construction, such as chaining subtasks, adding hops or following dependency graphs~\citep{xie2026agentsynth,shi2026taskcraft,fan2026scalable,deng2026swemilestone}, or check it with an LLM that reads the task or a trajectory~\citep{zou2026scaling,wu2026terminaluniverse}. Neither measures whether a later step requires the agent to read the output of an earlier one, so a step can add operations without requiring the agent to look at the environment. When we state a turn cap in the instruction and stop the agent once it is reached (turn-capped evaluation), most synthetic tasks pass within a few turns; in many of them the agent issues every command in its first turn and never reads any output, so the task reduces to single-turn generation. Solving these problems matters for both uses: a benchmark should test whether an agent can sustain a long interaction, and training data should contain such interactions.

\begin{figure}[t]
  \centering
  \includegraphics[width=\linewidth]{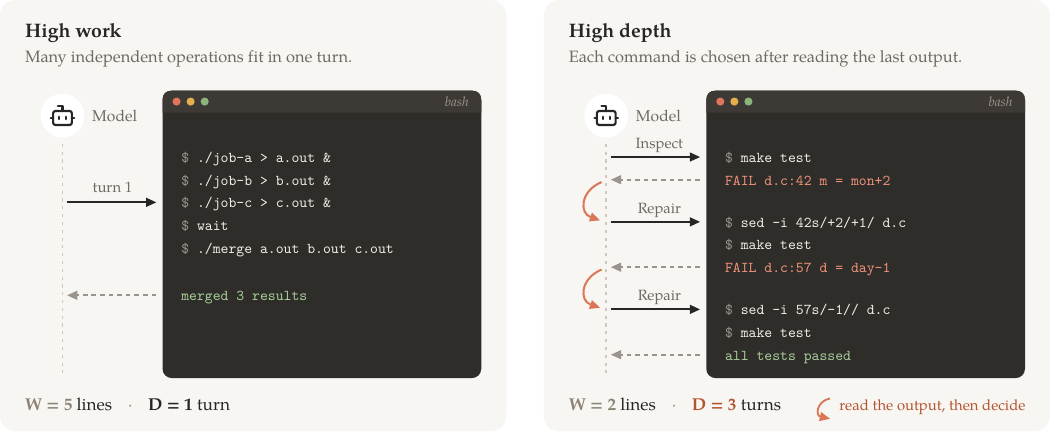}
  \caption{Work $W$ and depth $D$, as exchanges between the model and a terminal; time runs downward. High $W$: many independent operations that fit in one turn. High $D$: a chain of steps, each decided only after reading the output of the previous one. The commands are illustrative.%
  }
  \label{fig:concept}
\end{figure}

What makes a terminal task long? Many operations alone do not: a task whose operations are independent can be completed by one script in one turn.
A task is long when the agent must, again and again, read the output of one command before it can decide the next. In the human-curated tasks that take many turns under a cap, this need arises from details of real software that the agent does not know in advance, such as the default configuration of a specific version, the interface of an old release or a truncated file in a real format; in systems with several components, fixing one such problem often exposes the next. Configuration and version details are a common cause of failures in deployed systems~\citep{yin2011empirical}, and language models handle rarely documented and version-specific knowledge poorly~\citep{kandpal2023large,misra2026gitchameleon}. Task length thus has two separate dimensions, the amount of work and the number of turns that must happen in sequence, and the second largely comes from environments the agent cannot predict.

Based on this observation, we describe a terminal task by two quantities used in parallel computing, its work and its depth~\citep{blelloch1996programming}. As illustrated in \autoref{fig:concept}, the work $\Wstar$ is the number of operations needed to complete the task when the state of the environment is known; the depth $D^*$ is the number of turns that must happen in sequence, because the commands of each turn can only be chosen after reading the output of the previous one. We estimate $\Wstar$ by the number of lines $W$ of the reference solution, and $D^*$ by turn-capped evaluation: the smallest cap under which the task passes is recorded as $D$. With these estimates we audit six existing datasets and find that existing synthesis methods primarily increase work rather than depth.
We also find that tasks with both high work and high depth are substantially more challenging for current models.

To obtain tasks that are large in both, we propose \ours, in which a generation agent \textbf{D}erives \textbf{E}ach segment of a task by \textbf{P}erforming and \textbf{T}esting it in a real software environment, \textbf{H}anding the state it leaves to the next segment.
Each task consists of segments delivered one at a time in the same session, so later segments depend on what earlier ones changed. The resulting tasks exhibit greater work and depth than any of the six audited datasets, while providing meaningful discrimination across models with different capabilities. As shown in \autoref{tab:bench}, even frontier models achieve pass rates below 50\%, highlighting the difficulty of these tasks. Beyond evaluation, trajectories collected from these tasks also serve as training data: fine-tuning a small open model on them improves its performance on existing single-instruction terminal benchmarks~\citep{merrill2026terminalbench,tbench2026tb21,otagent2026tblite}.

Our contributions are as follows.
\begin{enumerate}[leftmargin=*,itemsep=1pt,topsep=2pt, nosep]
  \item \textbf{Measurement and audit.} We describe terminal-task length by work $\Wstar$ and depth $\Dstar$, estimate $D$ per task by turn-capped evaluation, and use it to audit six existing datasets, finding that synthesis methods mainly add work.
  \item \textbf{Construction and evaluation.} We build deep terminal tasks from real software environments by segmented generation, and show that they are substantially deeper than existing datasets and challenging for current models.
  \item \textbf{Training.} We show that fine-tuning Qwen3-8B~\citep{yang2025qwen3} on trajectories from our tasks improves its performance on two existing terminal benchmarks.
\end{enumerate}

We discuss related work in \autoref{app:related}.

\section{Preliminaries}
\label{sec:prelim}

\textbf{Harbor tasks.} A task $\tau=(I,e_0,\sigma,V)$ in the Harbor format~\citep{harbor2026} consists of an instruction $I$, a Docker environment with initial state $e_0$, a reference solution $\sigma$ (a shell script), and a verifier $V$. We use the multi-segment form supported by Harbor, in which the $k$ segments of a task, $I=(I_1,\dots,I_k)$ and likewise $\sigma$ and $V$, are delivered one at a time in one container. Each $V_s$ is a set of checks $c$ on the state $e$ of the environment, with $c(e)=1$ if the check passes, and segment $s$ passes on $e$ if
\begin{equation}
  V_s(e)=\prod_{c\in V_s}c(e)=1.
\end{equation}
The agent sees only $I$ and the container, never $\sigma$ or $V$.

\textbf{Turns and runs.} The agent acts through a harness that relays commands and outputs between the model and the container (terminus-2~\citep{merrill2026terminalbench} in this paper). In turn $t$, the agent writes one reply containing a batch of commands $a_t$, possibly a script with branches and loops; the harness executes $a_t$ and returns the terminal output $o_t$. The harness limits the length of a reply, truncates long outputs and stops waiting for a command after a time limit. A run $(a_1,o_1,\dots,a_n,o_n)$ has $n$ turns, counted over all segments. Let $\hat e_s$ be the state in which the agent ends segment $s$, with $V_s(\hat e_s)=0$ if the run stops earlier. A run passes if
\begin{equation}
  R=\prod_{s=1}^{k}V_s(\hat e_s)=1.
\end{equation}
In turn-capped evaluation with cap $T$, the agent is told that at most $T$ turns are available, and the harness stops the run after turn $T$, so $n\le T$.

\textbf{Work and depth.} A strategy $\pi$ chooses each batch from what the agent has seen, $a_t=\pi(I_1,\dots,I_s,o_1,\dots,o_{t-1})$ in segment $s$, with general knowledge built into $\pi$. Let $n_\pi(e)$ be the number of turns $\pi$ needs to pass from initial state $e$ ($\infty$ if it never passes), and let $\mathcal{E}(\tau)$ be the set of initial states that differ from $e_0$ only in details the instruction leaves open, such as file contents, versions and default settings, and that $\sigma$ solves once adapted to these details; before acting, the agent cannot tell them apart, as in a partially observable environment. The depth of $\tau$ is
\begin{equation}
  \Dstar=\min_{\pi}\,\max_{e\in\mathcal{E}(\tau)}\, n_\pi(e),
\end{equation}
the fewest turns that suffice whichever state in $\mathcal{E}(\tau)$ the agent faces. $\Dstar$ thus counts rounds of interaction between the model and the terminal, as adaptive complexity counts rounds of queries~\citep{balkanski2018adaptive}, rather than dependencies between operations: a dependency that a batch can resolve by branching on the output of its own commands adds no turn, and two turns can be merged only if the later commands, branches included, can be written within one reply without reading the earlier output. Merging is limited because a reply has bounded length and the model takes no part while a batch runs: a batch can branch only on outcomes the model lists in advance, and a step that depends on an output it could not anticipate, such as an unfamiliar error or code to be patched, needs a turn of its own. Turns spent reading long outputs or waiting for slow commands also count. Let $\mathcal{A}$ be the set of command sequences, written in any number of turns, that pass from $e_0$ when $e_0$ is known and the harness limits on reply length and waiting time are lifted; $\mathcal{A}$ contains $\sigma$.
The work of $\tau$ is
\begin{equation}
  \Wstar=\min_{a\in\mathcal{A}}\,|a|,
\end{equation}
where $|a|$ counts lines, one command per line and each line of file content written by $a$ counted separately, summed over segments. Unlike $\Dstar$, $\Wstar$ is taken at $e_0$ alone, so that the cost of not knowing the state falls entirely on $\Dstar$: when $e_0$ is known, %
nothing needs to be observed, so exploration and error recovery count toward $\Dstar$ but not $\Wstar$.

Given a set of models $\mathcal{M}$ and a set of caps $\mathcal{T}$, let $P_T=1$ if a run of a model in $\mathcal{M}$ passes from $e_0$ under cap $T$. We use $D=\min\{T\in\mathcal{T}:P_T=1\}$, the smallest cap under which a model in $\mathcal{M}$ passes in our evaluation protocol, as an empirical estimate of $\Dstar$. Unlike the idealized $\Dstar$, $D$ depends on the models, the harness and the single state $e_0$, so it is an estimate rather than a bound. We use the number $W$ of non-blank, non-comment lines of $\sigma$, summed over segments, as an estimate of $\Wstar$.

\section{Method}
\label{sec:method}

As shown in \autoref{fig:pipeline}, \ours has two parts. The first measures depth by turn-capped evaluation over a set of models; we apply it to existing datasets and to our own tasks. The second builds tasks whose depth comes from the environment rather than from how the instruction is written: a generation agent carries out each segment in a real software environment before writing it down. Runs on these tasks also serve as training data.

\begin{widefigure}[t]
  \centering
  \includegraphics[width=\textwidth]{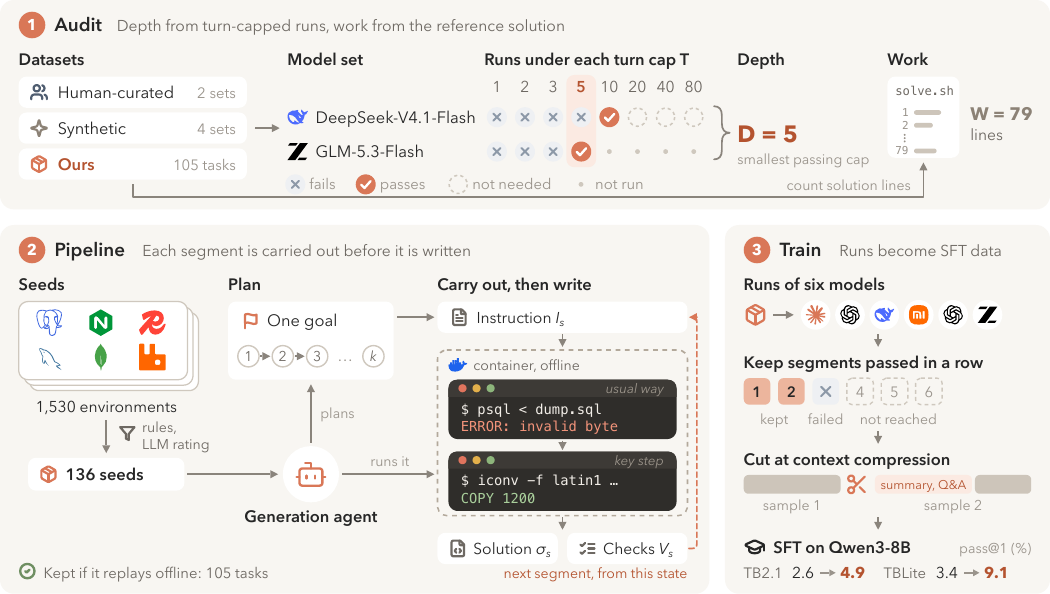}
  \caption{Overview of \ours. \textbf{Audit} measures depth $D$ by runs of two models under stated turn caps and work $W$ by the lines of the reference solution. \textbf{Pipeline} builds multi-segment tasks from real software environments: for each segment, the generation agent writes the instruction, carries the segment out in the container, and then writes the solution and the checks. \textbf{Train} keeps the segments a run passes in a row and cuts runs where the context is compressed. The turn caps, the solution length and the commands are illustrative.}
  \label{fig:pipeline}
\end{widefigure}

\subsection{Measuring Depth}
\label{sec:measure}

The number of turns a model takes under a loose limit reflects how the model works as much as what the task requires~\citep{meng2026calibforge,desai2026swemarathon}. We use a stated cap to measure the task: the agent is told the cap and stopped once it is reached, and the smallest cap under which the task passes serves as an estimate of $\Dstar$.

\textbf{Caps.} All datasets share the caps $\mathcal{T}=\{1,2,3,5,\allowbreak 10,20,\allowbreak 40,80\}$. For a model $m$, let $P^m_T=1$ if $m$ passes the task under cap $T$. The estimate from $m$ is
\begin{equation}
  D_m=\min\{T\in\mathcal{T}:P^m_T=1\}.
\end{equation}
A task that $m$ passes only without a cap is recorded as above the largest cap, and a task that $m$ never passes gives no estimate.

\textbf{Model set.} Once told the cap, models differ widely in how far they merge commands into fewer turns, and a model that merges poorly overestimates the depth. We therefore measure every task with a set of models $\mathcal{M}$ and take the smallest estimate,
\begin{equation}
  D=\min_{m\in\mathcal{M}} D_m,
\end{equation}
over the models that give one. In this paper, $\mathcal{M}$ consists of DeepSeek-V4.1-Flash~\citep{deepseekai2026deepseekv41flash} and GLM-5.3-Flash~\citep{zai2026glm53flash}. We show more details in \autoref{app:measure}.

\subsection{Seed Environments}
\label{sec:seeds}

We draw seed environments from public terminal-task packages, mostly the 1,530 packages of TerminalWorld~\citep{chu2026terminalworld}; each package comes with a software environment built from a Dockerfile. After filtering by rules, an LLM judge rates each environment on whether its components are real software rather than stand-ins, whether several of them work together, whether the software is uncommon, and whether it can support several segments of work. Keeping at most one environment for each kind of system, we collected 136 seed environments. Generation and evaluation both run without network access, so the generation agent meets the problems the evaluated agent will meet, and reference solutions can be replayed later.

\subsection{Segmented Generation}
\label{sec:gen}

When a whole task is generated in one pass, it is hard to ensure that the intermediate states reached while completing it depend on one another. We instead generate a task segment by segment inside the container: a generation agent, an LLM agent working in the seed environment, carries out each segment and writes the next one from the state it leaves.

\textbf{Carrying out before writing.} The agent first explores the environment and plans one overall goal that a user of this software would pursue, split into $k$ segments. For segment $s$, it writes $I_s$, carries out the segment in the container, and then writes $\sigma_s$ and $V_s$ from what the software actually did. $I_s$ states the goal, the deliverables that will be checked and the constraints, and refers to earlier results by name; it does not say how to do the work or what will go wrong. As generation proceeds, a clean container replays $e_s=\sigma_s(e_{s-1})$ for $s=1,\dots,k$, where $\sigma_s(e)$ is the state reached by running $\sigma_s$ from state $e$, so that $\sigma$ does not rely on changes the agent made by hand.

\textbf{Key steps from observed problems.} Each segment has at least one key step: a step where the usual way of doing it goes wrong in this environment, with an error or silently, and only running the commands reveals it. We write $\tilde\sigma_s$ for $\sigma_s$ with its key step removed.

\textbf{Dependence between segments.} Let $\mathrm{res}(e)$ be the results of segments, such as data, files, accounts or configuration, that hold in state $e$. Segment $s+1$ must use a result in
\begin{equation}
  R_s=\mathrm{res}(e_s)\setminus\mathrm{res}\bigl(\tilde\sigma_s(e_{s-1})\bigr),\qquad s<k,
\end{equation}
which exists only once the key step of segment $s$ is done; the evaluated agent thus has to read the output and change course at the key step. The last segment performs one operation on the whole system, such as moving it to a new data directory, after which earlier results must still hold, \mbox{$\mathrm{res}(e_{k-1})\subseteq\mathrm{res}(e_k)$}.

\subsection{Verifier Design and Acceptance}
\label{sec:verifier}

The verifier decides what counts as completing a segment, and thus which steps an agent cannot skip: weak tests accept invalid solutions~\citep{yu2025utboost}, and a key step that the verifier does not see adds no depth. The agent writes each $V_s$ under three rules, using a fixed library of comparison functions.

\textbf{Outcomes observed through the software.} A check asks the running system with its own tools: a client gets an answer from the service, the software's validator accepts the configuration, a query returns the right rows. Checks do not look for particular settings or commands, so any correct way of reaching the goal passes. The checks of the last segment, $V_k$, examine the results of all earlier segments again.

\textbf{Tied to the key step.} The agent writes $V_s$ so that it fails both on the state before segment $s$ and on the state that $\tilde\sigma_s$ leaves, that is, at least one of its checks fails in each case:
\begin{equation}
  V_s(e_{s-1})=0, \qquad V_s\bigl(\tilde\sigma_s(e_{s-1})\bigr)=0.
  \label{eq:tied}
\end{equation}
The first condition, common in task verification~\citep{jimenez2024swebench}, makes the checks test the work of segment $s$; the second makes them see its key step.

\textbf{Stated in the instruction.} Every name a check of $V_s$ relies on, such as a path, a port or an account, must appear verbatim in $I_s$, while the fixes of key steps must not. An agent therefore knows what will be checked but not how to get there.

\textbf{Acceptance.} A task is accepted if replaying $\sigma$ from $e_0$ in a fresh offline container passes every segment:
\begin{equation}
  V_s(e_s)=1, \qquad s=1,\dots,k.
\end{equation}
Acceptance rests on this replay. The conditions in \eqref{eq:tied} guide how the agent writes the checks, and we record how often they hold by replaying accepted tasks with parts of $\sigma$ removed, such as a key step. We show more details in \autoref{app:checklib}.

\subsection{Training Data}
\label{sec:train}

We turn runs on our tasks into data for supervised fine-tuning (SFT). We collect runs without a turn cap from the six models we evaluate, and from each run we keep the segments that pass in a row from the first one. These runs often exceed the context window of the model being trained, and at evaluation terminus-2 then compresses the context into a summary of the work so far and questions and answers about it. We therefore cut each run where terminus-2 would compress it and start the next sample from this summary and these questions and answers, so that the trained model also learns to continue after a compression. We fine-tune Qwen3-8B on the resulting samples.

\section{Experiments}
\label{sec:experiments}

We first audit existing datasets with $D$ and test whether $D$ tracks difficulty, then measure our tasks, evaluate six models on them, and fine-tune a small model on their runs.

\subsection{Setup}
\label{sec:exp-setup}

\textbf{Datasets.} We audit four synthetic datasets, TMax~\citep{ivison2026tmax}, FACET~\citep{shi2026facet}, SETA~\citep{shen2026seta} and RST~\citep{li2026rst}, and two human-curated ones, TB2~\citep{merrill2026terminalbench} and TW~\citep{chu2026terminalworld}, sampling 734 tasks in total. Due to the implementation of the measurement method, a task that no model passes yields no estimate of $D$. Tasks that need GPUs, large downloads or privileged containers are excluded. We show more details in \autoref{app:sampling}. Our dataset contains 105 tasks.

\textbf{Measuring $D$.} Every task is measured with the two models in $\mathcal{M}$, both running in terminus-2~\citep{harbor2026}; adding GLM-5.3-Flash lowers $D$ for 25\% of the existing tasks it was run on and 7\% of our solved tasks. We show more details in \autoref{app:modelset}. A task counts as solved if DeepSeek-V4.1-Flash passes at least one of two runs without a cap, and statistics of $D$ are taken over solved tasks. We also record $N$, the number of turns in a passing run of DeepSeek-V4.1-Flash without a cap, which is the kind of turn count existing work reports.

\textbf{Evaluated models.} We evaluate DeepSeek-V4.1-Flash, GLM-5.3-Flash, MiMo-V2.6-Flash~\citep{xiaomi2026mimov26}, GPT-6 Luna~\citep{openai2026gpt6astra}, GPT-6.1 Sol~\citep{openai2026gpt61sol} and Claude Sonnet 5.5~\citep{anthropic2026sonnet55} on our tasks, all in terminus-2. Each model runs every task twice without a cap, and a run stops at the first segment it fails.

\textbf{Metrics.} pass@1~\citep{chen2021evaluating} is the share of runs that pass, averaged within each task and then over tasks. The fraction of segments completed is the number of segments a run passes in a row divided by $k$.

\subsection{Auditing Existing Datasets}
\label{sec:exp-audit}

\begin{widefigure}[t]
  \centering
  \includegraphics[width=\textwidth]{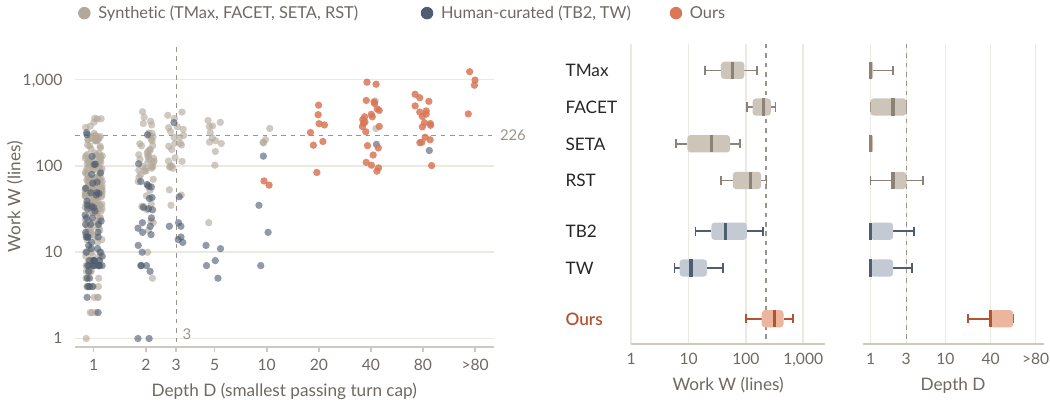}
  \caption{Depth $D$ and work $W$ of the solved tasks of seven datasets. Left: one dot per task, on log scales and jittered horizontally; grey dots are synthetic datasets, slate ones human-curated datasets and clay ones ours; $>80$ means passed only without a cap. Right: distributions of $W$ and $D$ per dataset; boxes span the 25th to 75th percentiles and whiskers the 10th to 90th. Dashed lines mark the 90th percentiles over the four existing synthetic datasets ($W=226$, $D=3$).}
  \label{fig:dw}
\end{widefigure}

\begin{widetable}[t]
  \caption{Audit of six existing datasets and ours. Solved: passed by DeepSeek-V4.1-Flash in at least one of two runs without a cap. Reported: length reported by the original paper (mean turns). $D\le3$, $D\ge5$: shares of solved tasks. $N$: fewest turns among passing runs without a cap. $D$, $N$ and $W$ are medians over solved tasks, and $\bar N$ is the mean of $N$. Minutes: time the agent runs in the container (median). Tokens: tokens (thousands) in the context at the end of the run (mean). Minutes and Tokens are from the run that gives $N$.}
  \label{tab:audit}
  \centering
  \small
  \setlength{\tabcolsep}{6.5pt}
  
  \begin{tabular}{lrrrrrrrrrrr}
    \headrow
    \hd{Dataset} & \hd{Sampled} & \hd{Solved} & \hd{Reported} & \hd{$D\le3$} & \hd{$D\ge5$} & \hd{$D$} & \hd{$N$} & \hd{$W$} & \hd{$\bar N$} & \hd{Minutes} & \hd{Tokens} \\
    \tabgroup{12}{Synthetic}
    TMax  & 100 & 94 & 16.3 & 100\% & 0\% & 1 & 5 & 58 & 5.8 & 0.9 & 6.6 \\
    FACET & 210 & 53 & 11.86 & 92\% & 8\% & 2 & 7 & 204 & 8.2 & 2.1 & 19.3 \\
    SETA  & 100 & 91 & -- & 98\% & 2\% & 1 & 5 & 25 & 7.0 & 0.8 & 8.4 \\
    RST   & 200 & 58 & -- & 79\% & 21\% & 2 & 7 & 119 & 8.2 & 1.2 & 11.0 \\[0pt]
    \tabgroup{12}{Human-curated}
    TB2   & 39 & 37 & -- & 89\% & 11\% & 1 & 8 & 44 & 10.9 & 2.2 & 16.3 \\
    TW    & 85 & 68 & -- & 90\% & 10\% & 1 & 9 & 11 & 11.6 & 1.5 & 14.4 \\[2pt]
    \oursrow
    \textbf{Ours}  & 105 & 54 & -- & 0\% & \best{100\%} & \best{40} & 45 & \best{315} & 52.8 & 12.7 & 67.0 \\
    \bottomrule
  \end{tabular}
\end{widetable}

\textbf{Synthetic datasets are mostly shallow.} \autoref{fig:dw} places the solved tasks of each dataset on the depth--work plane and shows how their $D$ and $W$ are distributed, and \autoref{tab:audit} summarizes each dataset. Of the 296 solved synthetic tasks, 278 (94\%) have $D\le3$. In TMax and SETA, nearly nine in ten tasks have $D=1$: a model passes them by issuing every command in the first turn, without reading any output.

\textbf{Turn counts overstate depth.} The number of turns a model takes without a cap is several times $D$. TMax and FACET report 16.3 and 11.86 turns on average, while the median $D$ of their solved tasks is 1 to 2 (\autoref{tab:audit}). On the 88 solved TMax tasks with Gemini runs released alongside the dataset, Gemini takes a median of 8 turns while the median $D$ is 1. Across the six datasets, 96 of the 170 tasks with $N$ between 6 and 10 have $D=1$.

\textbf{Synthesis adds work rather than depth.} The median $W$ of solved tasks is 68 lines for the synthetic datasets and 17 for the human-curated ones, yet the share with $D\ge5$ is no higher for the synthetic datasets (6\% versus 10\%). RST, which extends tasks over recursive rounds, has the deepest synthetic tasks and a median $W$ of 119 lines, and 79\% of its solved tasks still have $D\le3$. The human-curated sets hold a small deep tail: 11 of their 105 solved tasks have $D\ge5$, and runs on these tasks spend their turns waiting for slow commands, reading long outputs and recovering after version defaults, old command-line interfaces or truncated files mislead the model.

\subsection{Depth Predicts Difficulty}
\label{sec:exp-valid}

\begin{widefigure}[t]
  \centering
  \includegraphics[width=\textwidth]{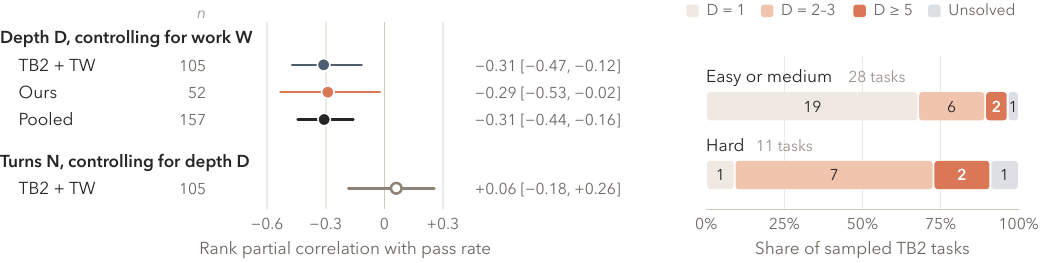}
  \caption{Depth against the pass rates of other models and the difficulty ratings of task authors. Left: rank partial correlation of the pass rate with $D$ controlling for $W$ (filled) and with $N$ controlling for $D$ (open), with 95\% CIs; pass rates are public ones for TB2 and TW and the mean of four evaluated models for ours. Right: distribution of $D$ among the sampled TB2 tasks, by the rating their authors gave (easy merged into medium); numbers are task counts.}
  \label{fig:dvalid}
\end{widefigure}

\textbf{Public pass rates.} TB2 and TW release per-task results of other models: the mean pass rate over 15 leaderboard submissions for TB2, and the share of eight models that solve each task for TW. We call these public pass rates. We relate them to $D$ by rank partial correlation~\citep{conover1981rank}, which replaces each variable by its rank and correlates two of them after removing the linear effect of a third, such as $W$.

\textbf{Depth predicts difficulty beyond work.} The left panel of \autoref{fig:dvalid} shows that, once $W$ is controlled, tasks with larger $D$ have lower public pass rates: the rank partial correlation is $-0.31$ on the 105 solved TB2 and TW tasks and $-0.49$ on TB2 alone. On our tasks, the same test uses the mean pass rate of GLM-5.3-Flash, GPT-6 Luna, MiMo-V2.6-Flash and GPT-6.1 Sol; because GLM-5.3-Flash is one of them, $D$ here comes from DeepSeek-V4.1-Flash alone. Over the 52 solved tasks with results from all four, the correlation is $-0.29$ (95\% CI $-0.53$ to $-0.02$), close to that of the human-curated sets, and $-0.31$ when the two are pooled.

\textbf{Turn counts do not predict difficulty once depth is known.} Controlling for $D$, $N$ is unrelated to the public pass rate on TB2 and TW ($+0.06$, 95\% CI $-0.18$ to $+0.26$; open circle in \autoref{fig:dvalid}).

\textbf{Difficulty labels from generation barely track pass rates.} As \autoref{fig:labels} shows, the labels set during generation correlate with the released pass rates at only $-0.07$ to $-0.11$, and those in the task files of TW not at all, while the ratings of TB2 authors reach $-0.46$.

\textbf{Depth agrees with the ratings of task authors.} Of the six datasets, only TB2 states that its difficulty labels are written by people: each task author rates the task as easy, medium or hard. Nine of the 11 sampled tasks rated hard have $D\ge2$, against 8 of the 28 rated easy or medium (right panel of \autoref{fig:dvalid}). On the 37 solved tasks, the rank partial correlation of the rating with $D$ controlling for $W$ is $+0.55$ (95\% CI $+0.29$ to $+0.75$), while those of $W$ and $N$ with the rating, each controlling for $D$, are not significant ($+0.16$ and $-0.28$, both CIs containing zero).

\textbf{Only tasks large in both are harder.} Splitting the TB2 and TW tasks at $D=2$ and at the median $W$, the tasks large in both have a mean within-dataset percentile of the public pass rate of 0.29 (lower is harder), against 0.53 to 0.59 for the other three groups; a task large in only one of the two is not harder.

\subsection{Our Tasks}
\label{sec:exp-ours}

\textbf{Our tasks are deep and large.} Each of the 105 tasks comes from a different seed environment and has three to six segments, six for most. DeepSeek-V4.1-Flash solves 54 of them, and 44 of these have $D\ge40$ (\autoref{fig:dw}). Their median $W$ is 315 lines, above the 90th percentile of $W$ over solved tasks of the existing synthetic datasets (226 lines), and their median $N$ is 45 turns, against 5 to 9 for the existing datasets (\autoref{tab:audit}).

\begin{figure}[t]
  \centering
  \includegraphics[width=0.8333\linewidth]{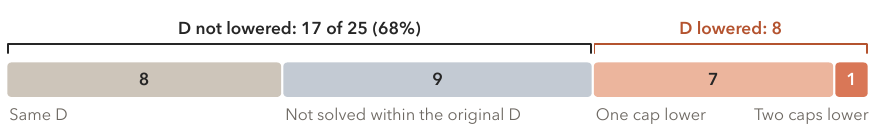}
  \caption{$D$ after the segments of a task are merged into one instruction, measured from the original $D$ downward. Numbers are task counts.}
  \label{fig:merge}
\end{figure}

\textbf{Depth does not come from delivering segments.} Each segment takes at least one turn, so $D$ is at least the number of segments, at most six, while 44 tasks have $D\ge40$. To remove this bound, we merge the segments of 25 randomly sampled tasks into one instruction and measure $D$ again. As shown in \autoref{fig:merge}, $D$ does not decrease for 68\% of them. Of the tasks whose $D$ drops, most drop by only one cap, suggesting that depth comes mainly from what the tasks require.

\textbf{Depth does not come from the seed environments.} Of the 54 seed environments of our solved tasks, 35 are released with an original task whose $D$ can be measured. Paired with our tasks from the same environments, these original tasks have $D\le10$ in 34 pairs and $D=1$ in 26, while ours have $D\ge20$ and the higher $D$ in all 35 (sign test, $p=5.8\times10^{-11}$).

\textbf{Checks see key steps and the dependence between segments.} When we replay reference solutions with parts removed, the checks of a segment fail before the segment is done in 601 of 602 segments (99.8\%) and fail when its key step is removed in 598 of 604 (99.0\%). With the key step removed, the next segment also fails in 421 of 500 cases (84.2\%), so later segments build on what the key step produces. We show more details in \autoref{app:checklib}.

\textbf{Cost.} Generation cost \$113.66 for the 136 seed environments, from which 105 tasks were accepted (77\%): about \$1.08 per accepted task, excluding evaluation and the measurement of $D$.

\subsection{Evaluating Models}
\label{sec:exp-bench}

\begin{figure}[t]
  \centering
  \includegraphics[width=\linewidth]{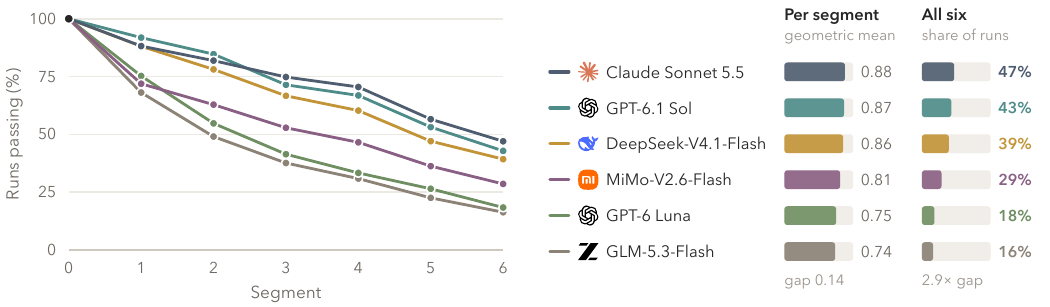}
  \caption{Runs segment by segment. Left: share of runs that pass segment $i$, among runs on tasks with at least $i$ segments. Right: the per-segment pass rate, the geometric mean of the conditional pass rates, next to the share of runs that pass all six segments, on one scale. Values are pooled over runs rather than averaged per task as in \autoref{tab:bench}.}
  \label{fig:survival}
\end{figure}

\begin{widetable}[t]
  \caption{Results of six models on our tasks. Effort: reasoning effort (--: not configurable). pass@1: share of passing runs, averaged within each task and then over tasks. Segments: mean fraction of segments passed. Minutes: time the agent runs in the container. Tokens: tokens (thousands) in the context at the end of a run.}
  \label{tab:bench}
  \centering
  \small
  \setlength{\tabcolsep}{3.9pt}
  
  \begin{tabular}{lcrrrrrrr}
    \headrow
    & & & & \multicolumn{3}{c}{\textit{All runs}} & \multicolumn{2}{c}{\textit{Passing runs}} \\[-1pt]
    \headrow
    \hd{Model} & \hd{Effort} & \hd{pass@1 (\%)} & \hd{Segments} & \hd{Turns} & \hd{Minutes} & \hd{Tokens (K)} & \hd{Turns} & \hd{Minutes} \\
    \tabgroup{9}{Proprietary}
    \tabmark{claude}Claude Sonnet 5.5   & high   & \best{47} & \best{0.70} & 36 & 10.4 & 40.3 & 38 & 11.1 \\
    \tabmark{openai}GPT-6.1 Sol         & medium & 44 & 0.69 & 32 & 5.7  & 29.0 & 34 & 6.1 \\
    \tabmark{openai}GPT-6 Luna          & medium & 20 & 0.43 & 35 & 6.9  & 32.5 & 38 & 7.0 \\
    \tabgroup{9}{Open-weight}
    \tabmark{deepseek}DeepSeek-V4.1-Flash & high   & 40 & 0.64 & 46 & 13.5 & 58.5 & 53 & 14.9 \\
    \tabmark{xiaomi}MiMo-V2.6-Flash     & --     & 30 & 0.50 & 48 & 39.5 & 50.0 & 62 & 45.8 \\
    \tabmark{zai}GLM-5.3-Flash          & max    & 17 & 0.38 & 26 & 24.6 & 37.3 & 43 & 32.6 \\
    \bottomrule
  \end{tabular}
\end{widetable}

\textbf{Our tasks separate models, and frontier models pass fewer than half.} In \autoref{tab:bench}, pass@1 ranges from 17\% for GLM-5.3-Flash to 47\% for Claude Sonnet 5.5, a spread of 30 points, with GPT-6.1 Sol at 44\%. Of the 90 tasks with results from all six models, 29 are passed by none.

\textbf{Long horizons amplify per-segment gaps.} \autoref{fig:survival} follows runs segment by segment. We summarize how reliably a model completes one segment by the geometric mean of its conditional pass rates, the share of runs reaching a segment that pass it, which ranges from 0.88 for Claude Sonnet 5.5 to 0.74 for GLM-5.3-Flash. A gap of at most 0.14 per segment compounds over six segments into a nearly threefold gap in the share of runs that pass them all (47\% versus 16\%), consistent with how small differences in per-step reliability compound over long tasks~\citep{sinha2026illusion}.

\subsection{Fine-Tuning}
\label{sec:exp-sft}

\textbf{Setup.} We fine-tune Qwen3-8B~\citep{yang2025qwen3} on 4,941 samples built from 1,272 runs on our tasks and evaluate it in terminus-2 on Terminal-Bench 2.1 (TB2.1;~\citealp{tbench2026tb21}) and OpenThoughts-TBLite (TBLite;~\citealp{otagent2026tblite}), with three runs per task. None of the tasks used for training is built from either benchmark. Both benchmarks give all requirements in one instruction. The baseline is the base model in the same non-thinking mode with the same sampling parameters.

\begin{figure}[t]
  \centering
  \includegraphics[width=0.5\linewidth]{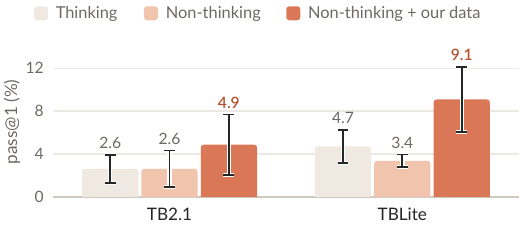}
  \caption{Qwen3-8B before and after fine-tuning on runs from our tasks. We fine-tune and evaluate in the non-thinking mode; the thinking mode of the base model is shown for reference. Bars are the mean pass@1 over three runs, and error bars one standard deviation.}
  \label{fig:sft}
\end{figure}

\textbf{Fine-tuning improves performance on both TB2.1 and TBLite.} As \autoref{fig:sft} shows, pass@1 rises from 2.6\% to 4.9\% on TB2.1 and from 3.4\% to 9.1\% on TBLite, and the fine-tuned model also exceeds the thinking mode of the base model on both. Runs on tasks delivered in segments thus carry over to tasks given in one instruction.

\section{Conclusion}
\label{sec:conclusion}

We presented \ours, which measures the depth of a terminal task, the number of turns that must happen in sequence, by turn-capped evaluation, and builds tasks that are large in both work and depth. The measurement shows that existing synthesis methods have mostly added work rather than depth, and that once work is controlled, deeper tasks are harder for other models. By having a generation agent carry out each segment in a real software environment and start the next segment from the state it leaves, \ours builds 105 tasks that are deeper and larger than those of every audited dataset; frontier models pass fewer than half of them, and fine-tuning a small open model on runs from these tasks improves it on two existing terminal benchmarks. These results suggest that the length of a task should be measured rather than assumed, and we recommend that builders of benchmarks and training data report depth alongside work.

\bibliography{refs}

\clearpage
\appendix
\makeatletter\def\fps@figure{!htb}\def\fps@table{!htb}\makeatother   %
\raggedbottom

\section{Related Work}
\label{app:related}

\textbf{Benchmarks for long-horizon agents.} Agents that act in computer environments are evaluated on resolving repository issues~\citep{jimenez2024swebench}, on terminal tasks~\citep{merrill2026terminalbench,chu2026terminalworld} and on security vulnerabilities, where SecProbe~\citep{luo2026secprobe} synthesizes repository-scale repair tasks on demand and chooses them adaptively with item response theory. Benchmarks that target long horizons report the length of a task as the time a human expert needs~\citep{kwa2025measuring} or as the turns or steps a model takes~\citep{li2026longhorizonterminalbench,desai2026swemarathon}, or make tasks long by chaining milestones on one codebase~\citep{deng2026swemilestone}. Related studies examine how reliability declines as tasks get longer~\citep{sinha2026illusion}. These measures describe what a model does under a loose limit. We instead state a turn cap and record the smallest cap under which a task passes, which estimates the turns that must happen in sequence rather than the turns a model happens to take.

\textbf{Synthesizing tasks for agents.} Training agents requires verifiable tasks and trajectories in large numbers~\citep{pi2026nemotron,raoof2026openthoughts,zou2026scaling}, and many pipelines synthesize them. Software-engineering tasks are built by injecting bugs into real repositories~\citep{yang2025swesmith}, and tool-use and computer-use tasks are generated automatically~\citep{shi2026taskcraft,xie2026agentsynth}. For the terminal, pipelines derive tasks from agent skills, skill graphs and agent trajectories~\citep{cheng2026terminalworld,fan2026scalable,wu2026terminaluniverse}, break environments~\citep{lin2026cligym}, and scale environments for reinforcement learning~\citep{peng2026litecoderterminal}; the synthetic datasets we audit, TMax~\citep{ivison2026tmax}, SETA~\citep{shen2026seta}, FACET~\citep{shi2026facet} and RST~\citep{li2026rst}, belong to this line. Synthetic trajectories also train components other than the agent, as in AuraGen~\citep{huang2026guardrail}, which injects risks of calibrated difficulty into agent trajectories to train a guardrail that checks plans before they are executed. Pipelines that aim for later steps to build on earlier ones obtain this by construction or check it with an LLM~\citep{xie2026agentsynth,shi2026taskcraft,wu2026terminaluniverse}. We instead let a generation agent carry out each segment in the container before writing the next, and measure the depth of the resulting tasks.

\textbf{Synthetic data generated by LLMs.} Beyond agent tasks, LLMs generate training and evaluation data across domains, and recent pipelines control what is generated and verify it before use. DataGen~\citep{huang2025datagen} is a unified framework for generating text datasets that uses attribute-guided generation for diversity, code-based checks for label correctness and retrieval for factual accuracy. ChemOrch~\citep{huang2025chemorch} synthesizes chemistry instructions with controlled diversity and difficulty and builds the responses with tools, repairing them from tool feedback. SpecAlign~\citep{wang2026specalign} turns model specifications written by providers into preference data through rule annotation and adversarial multi-agent synthesis. \citet{ma2026personalization} simulate multi-turn interactions among user, assistant and distractor agents to synthesize data for personalization, and SkillGen~\citep{ma2026skillgen} induces reusable skills from successful and failed agent trajectories and verifies each skill by comparing outcomes on the same instances with and without it. Our pipeline likewise verifies what it generates, but by execution in the environment itself: the reference solution is replayed in a fresh container, and the checks of each segment are written to fail when its key step is removed.

\section{Measurement Details}
\label{app:measure}

\textbf{Order of runs.} DeepSeek-V4.1-Flash first runs every task twice without a cap. If neither run passes, it gives no estimate for the task. Otherwise, let $N_m$ be the fewest turns among its passing runs; the $N$ reported in the main text is that of DeepSeek-V4.1-Flash. Caps at or above $N_m$ count as passed without being run, since a run has already passed within $N_m$ turns. The caps below $N_m$ are tried in increasing order with up to two runs each, the second run being skipped once the first passes, and the first cap with a passing run gives $D_m$. If no cap below $N_m$ passes, $D_m$ is the smallest cap at or above $N_m$, or above the largest cap if $N_m$ exceeds it.

\textbf{Second model.} GLM-5.3-Flash runs only at the caps below the estimate of DeepSeek-V4.1-Flash, with up to two runs per cap and no runs without a cap. It can therefore only lower $D$, and it is not run on tasks with $D=1$ or on tasks that DeepSeek-V4.1-Flash does not solve.

\textbf{Stating the cap.} Under cap $T$, the following text is appended to the instruction of an existing task:
\begin{casebox}{Turn limit appended to the instruction}
Turn limit: you have at most $T$ turns for this task. One turn is one of your responses; a response may contain several commands. The session is cut off after turn $T$ and whatever state the environment is in will be graded. Plan for this limit: batch commands where you can, and do not spend turns on steps that are not needed.
\end{casebox}
In multi-segment tasks the cap covers all segments together, the text says so (``for the whole task, all $k$ parts together'') and adds that parts not finished by turn $T$ fail, and every new segment reports the turns used so far. The harness stops the run after turn $T$.

\textbf{Harness.} Existing tasks and ours run in the same harness, terminus-2, and \autoref{tab:harness} compares its settings with those in the papers of the audited datasets, four of which also use terminus-2. After each turn, the agent sees the output produced since the previous turn, cut to 10,000 bytes by keeping the first and last 5,000. Commands do not block: for each command, the agent states how long to wait before the output is read, at most 60 seconds, and a command still running after that is waited for in a later turn. Both are defaults of terminus-2 and apply to all tasks alike. Without a cap, a run may take up to 150 turns. An existing task has 30 minutes, which replaces the agent time limit set by its dataset; this is longer than that limit for 477 of the 631 sampled tasks that set one, equal for 88 and shorter for 66, mostly SETA tasks with 60 minutes. Our tasks have 3 hours, or 2.5 and 1.5 hours for the 4 tasks with five segments and the 3 with three. All other parts of existing tasks are used as released.

\begin{table}
  \caption{Settings of the harness in the papers of the audited datasets and in ours. Time limit: agent time limit per task in the released tasks, averaged over all tasks of a dataset; for our tasks, that of the 98 with six segments. Output: the most output shown after a command, keeping its beginning and end. Wait: how long the harness waits before returning the output of a command; in terminus-2 the agent sets it, and in SETA a command still running after 20\,s is polled for up to 10\,s at a time. Turns: limit on the turns of a run. TMax and SETA give the settings of their RL rollouts; settings a paper does not state are the defaults of its released code. --: none reported.}
  \label{tab:harness}
  \centering
  \small
  \setlength{\tabcolsep}{5pt}
  \begin{tabularx}{\linewidth}{l l r l L l}
    \headrow
    \hd{Dataset} & \hd{Harness} & \hd{Time limit} & \hd{Output} & \hd{Wait per command} & \hd{Turns} \\
    \tabgroup{6}{Synthetic}
    TMax  & mini-SWE-agent based & -- & 10,000 characters & Blocks up to 120\,s & 64 \\
    FACET & terminus-2 & 2\,min & 10,000 bytes & Up to 60\,s, not blocking & No limit \\
    SETA  & CAMEL & 48\,min & 1,000 characters & Blocks up to 20\,s, then polls & 30 \\
    RST   & terminus-2 & 18\,min & 10,000 bytes & Up to 60\,s, not blocking & -- \\
    \tabgroup{6}{Human-curated}
    TB2   & terminus-2 & 28\,min & 10,000 bytes & Up to 60\,s, not blocking & No limit \\
    TW    & terminus-2 & 22\,min & 10,000 bytes & Up to 60\,s, not blocking & No limit \\
    \tabgroup{6}{This paper}
    Ours on existing tasks & terminus-2 & 30\,min & 10,000 bytes & Up to 60\,s, not blocking & 150 \\
    \oursrow
    \textbf{Ours} & terminus-2 & 3\,h & 10,000 bytes & Up to 60\,s, not blocking & 150 \\
    \bottomrule
  \end{tabularx}
\end{table}

\textbf{Delivering segments.} We extend terminus-2 to run the segments of a task in one conversation and one container; everything else, including its prompt templates and context compression, is unchanged. The first message announces the number of parts and explains the protocol:
\begin{casebox}{Start of the first message of a multi-segment task}
This task is handed to you in $k$ parts, one at a time, in this same terminal session. Below is part 1. When a part is done, set \verb|"task_complete": true|; the next part is then sent to you. Later parts build on what earlier parts left in the environment. Each part is graded when you mark it complete, and you are not told the grade; the task passes only if every part does.
\end{casebox}
When the agent marks a segment complete, its checks run, and the harness sends ``Part $i-1$ is marked complete. Here is part $i$ of $k$'' with the next instruction and the current terminal screen as a new message. Later instructions are never placed in the container, and the checks of a segment are removed from the container before the next segment starts, so the agent can read neither in advance. A run stops at the first segment that fails. terminus-2 asks the agent once to confirm that it is done; we keep this confirmation only in the last segment, since in earlier segments it would add one turn per segment and inflate $D$.

\textbf{Choice of models.} A model used to measure depth must merge commands into fewer turns once told the cap; otherwise it measures work. We test this on calibration tasks in which every step is solved by one short command. In the parallel tasks, ten independent steps are listed in one file. In the static-chain tasks, each of ten steps needs the answer to the previous one, but every step is written in a plain file from the start. Both can be completed in two or three turns by writing a script. In the control task, the instruction and data of each step are encrypted with the answer to the previous step, so every step needs a turn of its own. Under a cap of 3, DeepSeek-V4.1-Flash and GLM-5.3-Flash pass all 12 runs on the parallel and static-chain tasks, while DeepSeek-V4-Flash, which works through the steps one or two turns at a time, passes none. No model passes the six-step control under this cap. On three TMax tasks that DeepSeek-V4.1-Flash passes under a cap of 1, GLM-5.3-Flash also passes all three under that cap, so it writes working single-turn scripts on real tasks as well. Both models are also strong and inexpensive. On Terminal-Bench 2.1~\citep{tbench2026tb21}, DeepSeek-V4.1-Flash scores 90.6 and GLM-5.3-Flash 84.3 as reported by their developers~\citep{deepseekai2026deepseekv41flash,zai2026glm53flash}, against 88.8 for GPT-5.6 Sol reported by OpenAI~\citep{openai2026gpt56} and 89.1 for Claude Opus 5 in the report of DeepSeek-V4.1-Flash. Per million input and output tokens, DeepSeek-V4.1-Flash costs \$0.30 and \$1.20 and GLM-5.3-Flash \$0.15 and \$0.50~\citep{deepseek2026pricing,zai2026pricing}, against \$5 and \$30 for GPT-5.6 Sol and \$5 and \$25 for Claude Opus 5~\citep{anthropic2026opus5}.

\section{Dataset Selection and Sampling}
\label{app:sampling}

\textbf{Datasets.} We audit datasets of terminal tasks that release runnable task packages, each with an instruction, a Docker environment and a verifier. We evaluate four synthetic datasets: TMax, FACET, RST, and SETA. Among human-curated sets, TB2 consists of tasks written and verified by people, and TW is the subset of TerminalWorld that people reviewed, built from real terminal recordings.

\textbf{Sampling rules.} The same rules apply to every dataset.
\begin{enumerate}[leftmargin=*,itemsep=1pt,topsep=2pt]
  \item Only tasks that at least one model passes in the released results are drawn, since a task that no model passes gives no estimate of $D$; in the hardest level of TMax, only 1 of the 59 tasks with released Gemini runs was ever passed. FACET and part of SETA release no per-task results and are drawn at random.
  \item Tasks are split into groups by the length or difficulty the dataset reports (\autoref{tab:samplingrule}), drawn within each group with a fixed seed, and interleaved across groups, so that every prefix of the list is balanced across groups.
  \item Only tasks with light dependencies are kept: they run in an ordinary Docker container without GPUs, privileged mode or systemd, use no external services, and need neither downloads above 200\,MB nor builds above 1\,GB or 10 minutes. We decide by reading the instruction, Dockerfile, reference solution and verifier of each task, and build the image once when unsure.
  \item Once the image is built, the reference solution must get full marks and doing nothing must get zero; tasks that fail this check are replaced by the next ones in the list.
  \item Tasks are used as released, without changing their instructions, environments or verifiers.
\end{enumerate}
TB2 and TW are small, so all their tasks that meet the rules and pass the check are included. FACET and RST have low shares of solved tasks, so their lists are extended under the same rules until about 50 tasks are solved, which is why they have the most sampled tasks.

\begin{table}
  \caption{Groups and evidence of a passing model used to sample each dataset.}
  \label{tab:samplingrule}
  \centering
  \small
  \begin{tabularx}{\linewidth}{l L L r}
    \headrow
    \hd{Dataset} & \hd{Groups} & \hd{Evidence that a model passed} & \hd{Sampled} \\
    \tabgroup{4}{Synthetic}
    TMax  & Three complexity levels, each split into three groups by the fewest turns of a passing Gemini run; the hardest level as one group & Gemini-3-Flash passes at least 2 of 8 runs & 100 \\
    FACET & Three groups by number of checks & None released & 210 \\
    SETA  & Synth: hard level, drawn at random; Evolve: three groups by the fewest iterations of a passing Qwen3-30B run & Evolve: Qwen3-30B passes at least 2 of 8 runs in the authors' intermediate data; Synth: none & 100 \\
    RST   & Four groups by lines of the reference solution & Qwen3.5-27B runs, matched only to lineages (the tasks derived from one seed); lineages with no passed task are excluded & 200 \\
    \tabgroup{4}{Human-curated}
    TB2   & Passes of DeepSeek-V3.2 in its 5 leaderboard runs (5, 3--4, 1--2) & Per-run leaderboard results & 39 \\
    TW    & Empirical difficulty on the official site & Per-task results of DeepSeek-V4-Pro on the official site & 85 \\
    \bottomrule
  \end{tabularx}
\end{table}

\textbf{Sampling does not change the audit.} Sampling favors tasks that some model has passed and tasks with light dependencies, so a sample need not follow its original dataset. \autoref{tab:sampling} compares each sample with its original dataset over the labels the dataset releases, and reweights the share of solved tasks with $D\ge5$ to the original proportions of the labels. The synthetic samples lean toward harder labels, so reweighting lowers their share with $D\ge5$ (SETA from 2.2\% to 1.5\%, RST from 20.7\% to 14.9\%), while the shares of the human-curated sets rise by about one point. The hardest level of TMax is underrepresented because few of its tasks have ever been passed, and its 7 solved tasks all have $D\le3$.

\begin{table}
  \caption{Samples against the original datasets, by the labels each dataset releases. Reweighted: share of solved tasks with $D\ge5$ when each label takes its original proportion. $^\dagger$Over the labels present in the sample, which cover 42\% (SETA) and 78\% (TW) of the original tasks.}
  \label{tab:sampling}
  \centering
  \small
  \begin{tabularx}{\linewidth}{l P{3.3cm} L r}
    \headrow
    \hd{Dataset} & \hd{Labels} & \hd{Sample against original} & \hd{$D\ge5$: sample $\to$ reweighted} \\
    \tabgroup{4}{Synthetic}
    TMax  & Four complexity levels & First three about 30\% each (original 25\%); hardest 9\% (original 26\%) & 0\% $\to$ 0\% \\
    FACET & None (all ``medium'') & Three groups by number of checks, close to the original & 7.5\% (not reweighted) \\
    SETA  & Synth and Evolve, three levels each & Synth hard 52\% (original 14\%); Synth medium (original 57\%) not sampled & 2.2\% $\to$ 1.5\%$^\dagger$ \\
    RST   & Seed difficulty & Medium 27\% (original 18\%) & 20.7\% $\to$ 14.9\% \\
    \tabgroup{4}{Human-curated}
    TB2   & Author rating & Hard 28\% (original 34\%) & 10.8\% $\to$ 11.9\% \\
    TW    & Empirical difficulty & Very hard (original 22\%) not sampled & 10.3\% $\to$ 11.8\%$^\dagger$ \\
    \bottomrule
  \end{tabularx}
\end{table}

\section{Additional Audit Results}
\label{app:audit}

\begin{figure}
  \centering
  \includegraphics[width=\textwidth]{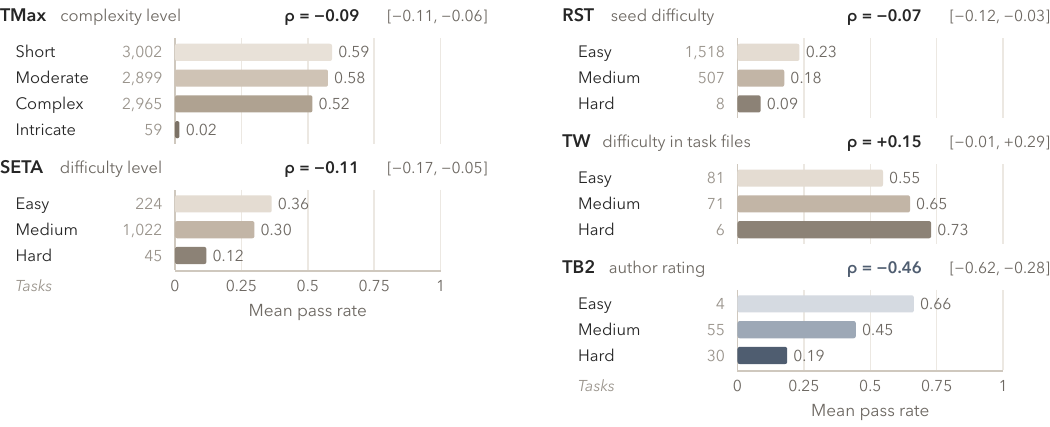}
  \caption{Mean pass rate at each difficulty level, from easiest to hardest, over all tasks with released pass rates (lineages for RST), with the number of tasks beside each level, and the rank correlation $\rho$ of the pass rate with the label. Labels: complexity level (TMax), difficulty level (SETA), seed difficulty (RST), difficulty in the task files (TW) and author rating (TB2).}
  \label{fig:labels}
\end{figure}

\textbf{Difficulty labels.} \autoref{fig:labels} gives the full comparison of difficulty labels with pass rates. It uses every task with a released pass rate rather than only the sampled ones: Gemini-3-Flash runs for TMax, runs of Qwen3-8B and Qwen3-30B for SETA, runs of Qwen3.5-27B for RST, released per lineage and counted once per lineage, the share of eight models on the TW website, and the mean over 15 leaderboard submissions for TB2. For TW we use the difficulty in the task files, since the difficulty on its website is computed from pass rates. In the synthetic datasets, the mean pass rate drops mainly at the hardest level, which holds few tasks.

\textbf{Shallow tasks test reasoning rather than observation.} A run that passes at cap $D$ issues about as many command lines as a passing run without a cap (1.06 times), while on tasks with $D=1$ the output at $T=1$ has 1.98 times as many tokens. The work stays the same, and the model replaces looking at the environment with reasoning about it.

\textbf{Runs on existing tasks stay short when unsolved tasks are included.} The statistics of $N$ in \autoref{tab:audit} are taken over solved tasks. \autoref{tab:runlength} also covers the tasks that DeepSeek-V4.1-Flash does not solve, averaging each sampled task over its two runs without a cap, whether or not they pass. Over all sampled tasks, runs take 6 to 13 turns, 1.0 to 3.3 minutes and 7 to 24 thousand tokens on the existing datasets, against 45 turns, 8.9 minutes and 57 thousand tokens on ours. On our tasks these values are lower than over solved tasks, because a run stops at the first segment it fails.

\begin{table}
  \caption{Runs of DeepSeek-V4.1-Flash without a cap. Solved tasks: $\bar N$, Minutes and Tokens as in \autoref{tab:audit}. All sampled tasks: each task is averaged over its two runs without a cap, whether or not they pass; Turns and Tokens are means over tasks and Minutes is the median.}
  \label{tab:runlength}
  \centering
  \small
  \begin{tabular}{lrrrrrr}
    \headrow
    & \multicolumn{3}{c}{\textit{Solved tasks}} & \multicolumn{3}{c}{\textit{All sampled tasks}} \\[-1pt]
    \headrow
    \hd{Dataset} & \hd{$\bar N$} & \hd{Minutes} & \hd{Tokens} & \hd{Turns} & \hd{Minutes} & \hd{Tokens} \\
    \tabgroup{7}{Synthetic}
    TMax   & 5.8 & 0.9 & 6.6 & 6.4 & 1.0 & 6.9 \\
    FACET  & 8.2 & 2.1 & 19.3 & 10.1 & 3.3 & 24.2 \\
    SETA   & 7.0 & 0.8 & 8.4 & 8.3 & 1.0 & 10.1 \\
    RST    & 8.2 & 1.2 & 11.0 & 12.0 & 2.1 & 15.2 \\
    \tabgroup{7}{Human-curated}
    TB2    & 10.9 & 2.2 & 16.3 & 13.4 & 2.6 & 19.2 \\
    TW     & 11.6 & 1.5 & 14.4 & 12.7 & 1.6 & 14.9 \\[2pt]
    \oursrow
    \textbf{Ours}  & 52.8 & 12.7 & 67.0 & 44.7 & 8.9 & 57.1 \\
    \bottomrule
  \end{tabular}
\end{table}

\begin{figure}
  \centering
  \includegraphics[width=\textwidth]{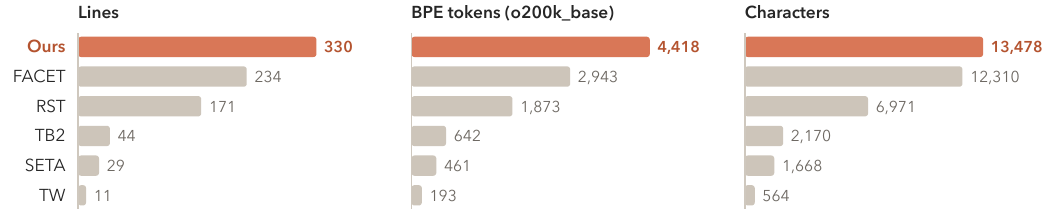}
  \caption{Median work of all sampled tasks, counted in lines, BPE tokens and characters of the reference solution. TMax releases no reference solutions.}
  \label{fig:wproxy}
\end{figure}

\textbf{Other measures of work.} $W$ counts the non-blank, non-comment lines of the reference solution, and lines depend on coding style. We therefore recount the same scripts in BPE tokens (the \texttt{o200k\_base} tokenizer of tiktoken) and in characters, summing over segments for our tasks as for $W$. TMax releases no reference solutions, so its $W$ instead counts the non-blank lines of the commands, file content included, that DeepSeek-V4.1-Flash sends in its passing run at the smallest cap it passes. TMax is left out here, which leaves 739 tasks, solved or not. Over these tasks, the Spearman correlation is 0.98 between lines and tokens and 0.97 between lines and characters. \autoref{fig:wproxy} shows the medians: the six datasets fall in the same order under all three measures, and ours is the largest under each, above FACET by 41\% in lines, 50\% in tokens and 9\% in characters.

\section{Robustness of $D$}
\label{app:modelset}

\subsection{The Model Set}

GLM-5.3-Flash runs only at the caps below the estimate of DeepSeek-V4.1-Flash, so its own estimate is known only on the tasks where it is lower. We therefore compare $D$ with the estimate of DeepSeek-V4.1-Flash alone.

\begin{figure}
  \centering
  \includegraphics[width=\textwidth]{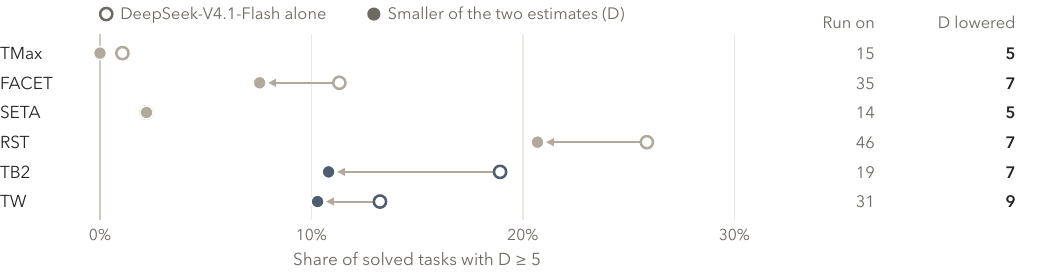}
  \caption{Share of solved tasks with $D\ge5$ in each existing dataset when $D$ comes from DeepSeek-V4.1-Flash alone and from the smaller of its estimate and that of GLM-5.3-Flash. Run on: solved tasks with $D>1$ on which GLM-5.3-Flash was run below the estimate of DeepSeek-V4.1-Flash. D lowered: those whose $D$ it lowers. In SETA the two shares coincide.}
  \label{fig:dshift}
\end{figure}

\textbf{Effect of the second model.} \autoref{fig:dshift} shows, for each existing dataset, the share of solved tasks with $D\ge5$ before and after taking the smaller estimate, together with the solved tasks on which GLM-5.3-Flash was run and those whose $D$ it lowers. The share falls in five of the six existing datasets and is unchanged in SETA. GLM-5.3-Flash lowers $D$ on 40 of the 160 existing tasks it was run on, 32 of them by one cap, and the two estimates order tasks alike: the Spearman correlation of $D$ with the estimate of DeepSeek-V4.1-Flash alone is 0.91 over the 401 solved existing tasks and between 0.79 and 0.93 within each dataset. The change concentrates in the deep tail of the existing datasets: of the 40 solved existing tasks whose estimate from DeepSeek-V4.1-Flash alone is at least 5, GLM-5.3-Flash brings 11 below 5. On our tasks, it was run on all 54 solved tasks and lowers $D$ on 4, each by one cap and none below 20, and the Spearman correlation is 0.95. Measuring with DeepSeek-V4.1-Flash alone would thus make the existing datasets look deeper, not shallower.

\begin{figure}
  \centering
  \includegraphics[width=\textwidth]{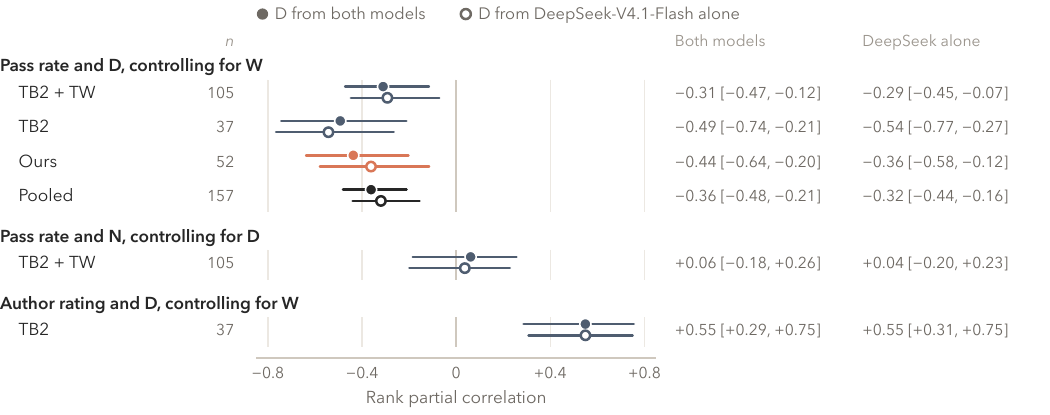}
  \caption{Rank partial correlations of the tests of depth against difficulty, with $D$ from both models (filled) and from DeepSeek-V4.1-Flash alone (open), with the number of tasks. For ours and the pooled rows, the pass rate is the mean of GPT-6 Luna, MiMo-V2.6-Flash and GPT-6.1 Sol, leaving out GLM-5.3-Flash, which takes part in measuring $D$.}
  \label{fig:modelset}
\end{figure}

\textbf{Results with DeepSeek-V4.1-Flash alone.} \autoref{fig:modelset} repeats the tests of depth against difficulty with each estimate. With DeepSeek-V4.1-Flash alone, the rank partial correlation of the public pass rate with $D$ controlling for $W$ is $-0.29$ on TB2 and TW and $-0.54$ on TB2, against $-0.31$ and $-0.49$ with both models; $N$ controlling for $D$ stays unrelated to the pass rate ($+0.04$); and the ratings of TB2 authors keep a correlation of $+0.55$ with $D$. On our tasks, the main text uses DeepSeek-V4.1-Flash alone, since GLM-5.3-Flash is one of the four models whose pass rates are averaged. With GLM-5.3-Flash left out of the average, the correlation is $-0.36$ with DeepSeek-V4.1-Flash alone and $-0.44$ with both models. The audit also changes little: with DeepSeek-V4.1-Flash alone, 92\% of the solved synthetic tasks have $D\le3$ (94\% with both models), and the TB2 and TW tasks large in both $D$ and $W$ remain the hardest group, with a mean within-dataset percentile of the public pass rate of 0.34 against 0.54 to 0.57 for the other three groups.

\subsection{Failed Runs}

$D$ is the smallest cap at which at least one of two runs passes. When DeepSeek-V4.1-Flash rarely passes a task, both runs at a sufficient cap can fail and raise $D$ by a cap; such tasks may also be harder for other models, so this error could produce a correlation between $D$ and difficulty that does not come from depth. Let $R$ be the share of the two runs of DeepSeek-V4.1-Flash without a cap that pass, which is 1 or 0.5 on solved tasks. We repeat the analyses on tasks with $R=1$, where a sufficient cap rarely fails twice. We do not control for $R$ instead: a deep task has many steps that can each go wrong, so part of the variation in $R$ may come from depth itself.

\begin{figure}
  \centering
  \includegraphics[width=\textwidth]{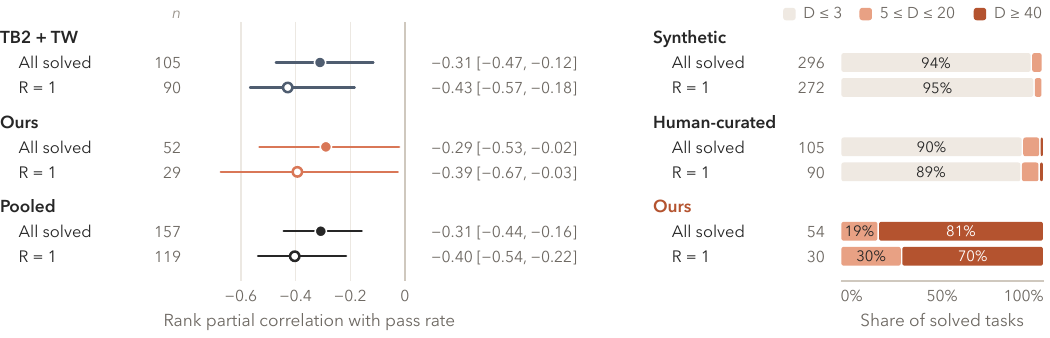}
  \caption{Robustness of $D$ to failed runs, over all solved tasks and over those with $R=1$, with the number of tasks beside each row. Left: rank partial correlation of the pass rate with $D$ controlling for $W$, computed as in \autoref{fig:dvalid}. Right: distribution of $D$, with $D$ as in \autoref{tab:audit}.}
  \label{fig:stability}
\end{figure}

\textbf{Depth still predicts difficulty.} On tasks with $R=1$, the rank partial correlations of the pass rate with $D$ controlling for $W$ are $-0.43$, $-0.39$ and $-0.40$ for the human-curated sets, ours and both pooled. All three are larger in magnitude than over all solved tasks (left panel of \autoref{fig:stability}).

\textbf{The depth distributions do not change.} On tasks with $R=1$, 95\% of the synthetic tasks have $D\le3$, and all of ours have $D\ge5$, 70\% of them $D\ge40$ (right panel of \autoref{fig:stability}).

\section{Checks}
\label{app:checklib}

A check script is a bash file in which every check prints one line, \texttt{PASS name} or \texttt{FAIL name: reason}; the segment passes if every check passes, and a script that stops before its last line fails. The generation agent writes checks with the functions in \autoref{tab:checklib}, which also do the comparing, so the agent decides only what to ask the system and what the answer must be.

\begin{table}
  \caption{Functions of the check library. \texttt{NAME} is a short check name, unique within the script.}
  \label{tab:checklib}
  \centering
  \small
  \begin{tabularx}{\linewidth}{l L}
    \headrow
    \hd{Function} & \hd{Passes when} \\
    \texttt{expect\_ok NAME -- CMD} & \texttt{CMD} exits with status 0 \\
    \texttt{expect\_out NAME PATTERN -- CMD} & a line of the output of \texttt{CMD} matches \texttt{PATTERN}, whatever the exit status \\
    \texttt{expect\_no\_out NAME PATTERN -- CMD} & \texttt{CMD} exits with status 0 and no line of its output matches \texttt{PATTERN} \\
    \texttt{expect\_fail NAME PATTERN -- CMD} & \texttt{CMD} fails and its output matches \texttt{PATTERN}, for an access that must be refused \\
    \texttt{expect\_eq NAME ACTUAL EXPECTED [KIND]} & the two values are equal as \texttt{KIND}: \texttt{text} (default), \texttt{int}, \texttt{num}, \texttt{version}, \texttt{bool}, \texttt{set}, \texttt{seq} or \texttt{exact} \\
    \texttt{expect\_file NAME PATH} & \texttt{PATH} is a non-empty file or a directory \\
    \texttt{expect\_port NAME HOST PORT} & a TCP connection to \texttt{HOST:PORT} succeeds \\
    \texttt{expect\_proc NAME PROGRAM} & a process runs \texttt{PROGRAM}, started by any path, with any arguments, directly or through an interpreter \\
    \texttt{expect\_proc\_arg NAME PROGRAM ARG\ldots} & such a process has every \texttt{ARG} on its command line \\
    \texttt{pass NAME}, \texttt{fail NAME REASON} & records a result the script decides itself \\[2pt]
    \headrow
    \hd{Reader} & \hd{Prints} \\
    \texttt{field FILE KEY} & the value of \texttt{KEY} in a report file of \texttt{name: value} lines or JSON \\
    \texttt{column FILE NAME:POS} & a column of a table, one value per line \\
    \texttt{cell FILE KEYNAME:POS KEYVALUE NAME:POS} & column \texttt{NAME} of the row whose column \texttt{KEYNAME} is \texttt{KEYVALUE} \\
    \texttt{rows FILE NAME:POS} & the number of data rows \\
    \texttt{proc\_start PROGRAM} & the start time of the newest process running \texttt{PROGRAM}, for events such as restarts \\
    \bottomrule
  \end{tabularx}
\end{table}

\textbf{Comparison.} Letter case, runs of spaces, and spaces around \texttt{:} and \texttt{=} are ignored in patterns and outputs; key and column names ignore case and treat \texttt{-}, \texttt{\_} and spaces alike; values lose surrounding spaces and quotes. The kinds \texttt{int}, \texttt{num} and \texttt{version} compare the first number or dotted version in each value, so \texttt{5.7.21-log} equals \texttt{5.7.21} as a version. \texttt{bool} maps yes, true, pass, ok, on and 1 to true and their opposites to false. \texttt{set} compares the items of two lists regardless of order and repeats, and \texttt{seq} compares them in order, for requirements such as ``X before Y''. An empty expected value fails, except that two empty sets are equal. Readers accept a missing header line, tabs, commas, \texttt{|} tables and aligned columns. Every command runs under a 120-second limit, and a check script must finish within 10 minutes.

\textbf{Rules for writing checks.} Besides the three rules in the main text, the generation agent follows these:
\begin{itemize}[leftmargin=*,itemsep=1pt,topsep=2pt]
  \item Expected values are computed from the live system when the checks run; counts, identifiers, versions or states seen during generation are never written into a check.
  \item Report files are read only through the readers, by the field and column names the agent declares; the pipeline writes the exact format into the instruction.
  \item An event the goal asks for, such as a restart or a failover, is checked through a trace the software itself keeps, such as a later start time than a component that was not restarted or a line in its own log. A file the agent writes about the event is not evidence, and an event that leaves no trace cannot be a deliverable.
  \item A string looked for in the files of the solver must appear in the instruction or be printed by the software itself, not by a script the agent wrote.
  \item A running service is checked by talking to it with its own client; processes are checked only when the instruction asks for a process.
  \item When the message of a tool and the real state disagree, or two readings of a requirement are defensible, both are accepted.
  \item Values that must be measured never appear among the names stated in the instruction, so checks cannot depend on them.
\end{itemize}

\textbf{Replays with parts removed.} The reference solution marks the commands of the key step of each segment, and those of events the goal asks for, with comments, so that they can be removed mechanically. For each of the 105 tasks, we replay the reference solution in a fresh offline container with one part changed and record whether the checks of the affected segment report a failure; a check script that crashes instead does not count. \autoref{tab:replay} lists the four changes. Segments whose replay did not run to a result, 2 to 15 for each change, are left out, and changes that involve the previous or next segment leave out the first or last segment of each task. These replays are not conditions for accepting a task and are not run during generation, so the generation agent never sees their results.

\begin{table}
  \caption{Replays of the reference solutions of all 105 tasks with one part changed. Holds: the checks of the segment named in the second column report a failure. Counts are over segments, except for removed events, which are counted one event at a time.}
  \label{tab:replay}
  \centering
  \small
  \setlength{\tabcolsep}{8pt}
  \begin{tabular}{lcrr}
    \headrow
    \hd{Change} & \hd{Checks expected to fail} & \hd{Holds} & \hd{Share} \\
    Checks of segment $s$ run before its work, on $e_{s-1}$ & $V_s$ & 601 / 602 & 99.8\% \\
    Previous segment skipped: $\sigma_s$ run without $\sigma_{s-1}$ & $V_s$ & 462 / 501 & 92.2\% \\
    One event of segment $s$ removed from $\sigma_s$ & $V_s$ & 158 / 165 & 95.8\% \\
    Key step of segment $s$ removed: $\tilde\sigma_s$ run instead of $\sigma_s$ & $V_s$ & 598 / 604 & 99.0\% \\
    Key step of segment $s$ removed, then $\sigma_{s+1}$ run & $V_{s+1}$ & 421 / 500 & 84.2\% \\
    \bottomrule
  \end{tabular}
\end{table}

\section{Our Dataset}
\label{app:data}

\textbf{Statistics.} \autoref{tab:ourdata} summarizes the 105 tasks. 98 have six segments, 4 have five and 3 have three. Instructions are longer after the first segment (median 244 words against 196), since they refer to earlier results by name and state which of them must be preserved. Of the 54 solved tasks, 2 have $D=10$, 8 have $D=20$, 23 have $D=40$ and 17 have $D=80$, and 4 are not passed within 80 turns. By the labels of their original packages, the seed environments cover software engineering (18 tasks), environment setup (17), system administration (16), security (15) and 18 other categories (35 tasks); 4 carry no label.

\begin{table}
  \caption{Statistics of our 105 tasks: median, with the 10th and 90th percentiles in brackets. Names: paths, ports, accounts and other names that the checks rely on, each stated verbatim in the instruction.}
  \label{tab:ourdata}
  \centering
  \small
  \setlength{\tabcolsep}{10pt}
  \begin{tabular}{lrr}
    \headrow
    & \hd{Per segment} & \hd{Per task} \\
    Lines of the reference solution & 52 [18, 130] & 330 [134, 626] \\
    Words of the instruction & 234 [171, 336] & 1,406 [1,153, 1,790] \\
    Names stated in the instruction & 14 [6, 24] & 83 [50, 123] \\
    \bottomrule
  \end{tabular}
\end{table}

\begin{figure}
  \centering
  \includegraphics[width=\textwidth]{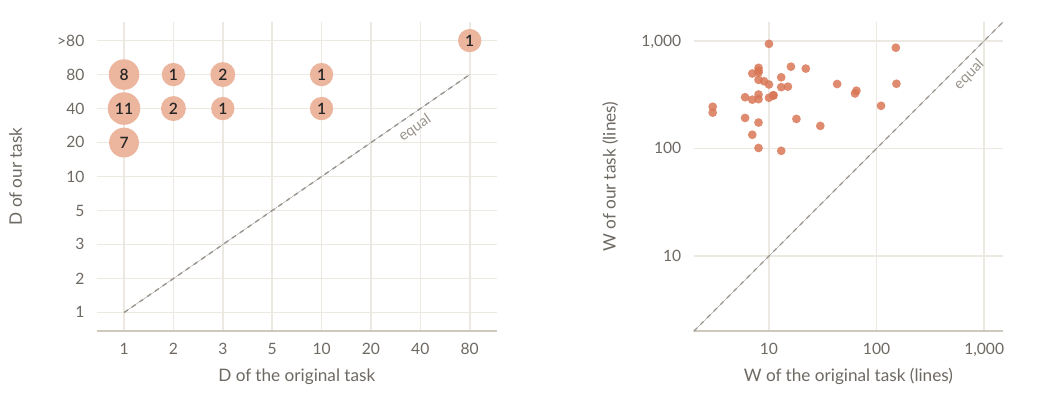}
  \caption{Original task of each seed environment against our task from the same environment, over the 35 pairs. Left: $D$, with the number of pairs in each cell; $>80$ means passed only without a cap. Right: $W$ in lines of the reference solution, on log scales. Dashed lines mark equal values.}
  \label{fig:seedpair}
\end{figure}

\textbf{Original tasks of the seed environments.} The reference solution of our task is longer than that of the original task in all 35 pairs, with a median of 324 lines against 10, and at least twice as long in each pair (right panel of \autoref{fig:seedpair}).

\textbf{An example task.} \autoref{tab:example} outlines a task built from an environment that holds the source of NetworkManager-l2tp 1.20.4, a VPN plugin for NetworkManager. Its $D$ is 40, $N$ is 37 and $W$ is 162 lines. Each segment works on the result of the previous one: segment 3 needs the plugin layout found in segment 2, segment 4 packages the corrected tree, and segment 6 relocates the package built in segment 5. GPT-6.1 Sol passes both of its runs, DeepSeek-V4.1-Flash and MiMo-V2.6-Flash one of two, and the other three models none. Below the table we show the instruction of segment 3 and part of its checks.

\begin{table}
  \caption{Segments of the example task and their key steps.}
  \label{tab:example}
  \centering
  \small
  \begin{tabularx}{\linewidth}{c P{5.6cm} L}
    \headrow
    \hd{Segment} & \hd{The instruction asks to} & \hd{Key step} \\
    1 & Generate the build system of the source checkout and configure it for a given installation layout & The \texttt{NetworkManager} pkg-config module is absent; the configure script actually needs \texttt{libnm}, which is installed \\
    2 & Build and stage the installation under \texttt{/app/nm-l2tp-stage}, leaving \texttt{/usr} untouched & The NetworkManager plugin libraries are staged under the multiarch library directory, not next to the service executable \\
    3 & Make the staged pppd plugin loadable by the installed pppd 2.4.9 while keeping the requested 2.4.7 layout, and write a manifest of the staged plugins & Copying the plugin into the 2.4.9 directory succeeds, but pppd looks up a bare plugin name in its own plugin directory; only loading it by its staged path shows that it works \\
    4 & Package the staged tree into one archive that can be unpacked elsewhere & The staged files record absolute paths of the stage, so a plain archive unpacks without error but works elsewhere only after these paths are made relative \\
    5 & Build a local Debian package from the archived payload & \texttt{dpkg-deb} rejects the tree until it has a control file, valid installation paths and executable permissions \\
    6 & Relocate the deployment to \texttt{/opt/nm-l2tp}, keeping all earlier results & The relocated service file still points to paths under \texttt{/usr} \\
    \bottomrule
  \end{tabularx}
\end{table}

\begin{casebox}{Instruction of segment 3}
\raggedright
\textbf{Goal.} The staged installation under \texttt{/app/nm-l2tp-stage} contains the NetworkManager-l2tp plugins prepared in earlier work. The host has pppd 2.4.9 installed, so the staged pppd plugin layout must match the pppd version that will load it while preserving the original 2.4.7 layout the user requested. The user also needs a manifest inventorying the staged shared-object plugins by path and checksum.

\textbf{Deliverables.}
\begin{itemize}[leftmargin=*,itemsep=0pt,topsep=1pt]
  \item The compiled pppd plugin shared object remains available under \texttt{/app/nm-l2tp-stage/usr/lib/pppd/2.4.7}.
  \item The staged pppd plugin is also present in the version-correct directory for the installed pppd, \texttt{/app/nm-l2tp-stage/usr/lib/pppd/2.4.9}, and the installed pppd is able to load it from there.
  \item The staged NetworkManager property plugin shared objects remain present in the build configured library plugin directory from \texttt{/app/nm-l2tp-stage}.
  \item \texttt{/app/nm-l2tp-stage/install-manifest.txt} records each staged shared-object plugin's name, resolved staged path, and SHA-256 checksum.
\end{itemize}

\textbf{Report files.} \texttt{/app/nm-l2tp-stage/install-manifest.txt}: a table with columns separated by a single tab character. The first line is the header \texttt{artifact path sha256}; then one line per plugin, in any order.

\textbf{Constraints.} Work only with the existing source checkout and staged installation from earlier segments. Preserve the original \texttt{/app/nm-l2tp-stage/usr/lib/pppd/2.4.7} plugin files. Do not install files into the live \texttt{/usr} tree. Do not modify \texttt{/opt/network-manager-l2tp.git} or the configured build source.
\end{casebox}

\begin{casebox}{Part of the checks of segment 3}
\footnotesize
\begin{verbatim}
STAGE=/app/nm-l2tp-stage
PPPD_DIR_249="$STAGE/usr/lib/pppd/2.4.9"
PPPD_PLUGIN=nm-l2tp-pppd-plugin.so
MANIFEST="$STAGE/install-manifest.txt"

expect_file staged-pppd-plugin-249 "$PPPD_DIR_249/$PPPD_PLUGIN"
pppd_load_249() {
    timeout 6 /usr/sbin/pppd plugin "$PPPD_DIR_249/$PPPD_PLUGIN" \
        nodetach noauth notty debug 2>&1 || true
}
expect_out pppd-loads-staged-plugin \
    "Plugin $PPPD_DIR_249/$PPPD_PLUGIN loaded[.]" -- pppd_load_249

live_so_paths() {
    find "$STAGE" -type f -name '*.so' -exec readlink -f -- {} \; | sort
}
expect_eq manifest-paths "$(column "$MANIFEST" path:2 | sort)" \
    "$(live_so_paths)" set
\end{verbatim}
\end{casebox}

\section{Merging Segments into One Instruction}
\label{app:merge}

We draw 25 tasks at random from the solved tasks whose $D$ had been measured when we ran this experiment and turn each into a task with a single segment. The environment and the work stay the same: the instruction gives the requirements of all segments at once in their original order, the reference solution runs the scripts of all segments in order, and the checks of all segments run together on the final state. Checks of earlier segments that a later segment is meant to change, such as a port that the last segment moves, are found by running the merged reference solution and dropped. A merged task passes the quality check if the reference solution then passes all remaining checks, including every check of the last segment.

We measure the merged tasks with DeepSeek-V4.1-Flash alone and without runs without a cap. Only caps up to the original $D$ of DeepSeek-V4.1-Flash are tried, from the largest downward, with two runs per cap, stopping at the first cap where both runs fail. The $D$ of a merged task is the smallest cap passed, and a merged task that fails at the original $D$ is counted as not solved within it.

For 17 of the 25 tasks, $D$ does not decrease: 8 pass at the original cap and 9 are not solved within it. Of the remaining 8, 7 drop by one cap level and 1 by two.

\section{Evaluation Details}
\label{app:eval}

\textbf{Setup.} All models run in the multi-segment version of terminus-2 with the limits used to measure $D$: 150 turns and the time limit of the task. The open-weight models and GPT-6 Luna are called through OpenRouter with the serving provider fixed: DeepSeek, Together or Wafer for DeepSeek-V4.1-Flash, Fireworks for GLM-5.3-Flash, Xiaomi for MiMo-V2.6-Flash and OpenAI for GPT-6 Luna. GPT-6.1 Sol and Claude Sonnet 5.5 are called through TokenRouter. We set the reasoning effort only for GPT-6 Luna (medium) and keep the default of the provider for the other models, which gives the levels in \autoref{tab:bench}. The runs of DeepSeek-V4.1-Flash for evaluation are separate from those for measuring $D$: it passes 52 tasks in its two evaluation runs, all among the 54 it solves when $D$ is measured.

\textbf{Refusals.} The provider of GPT-6.1 Sol refused requests on 7 tasks as a possible cybersecurity risk, and retries were refused as well. On 12 tasks, the safety classifier of Anthropic refused requests from Claude Sonnet 5.5~\citep{anthropic2026sonnet55}, returning a refusal without content and ending the run. On 3 more tasks Claude Sonnet 5.5 returned empty responses and gave no valid result. The 7 tasks refused for GPT-6.1 Sol are all among these 15. The refused tasks mostly concern system security, although every task runs offline. A task without a valid result is left out of the denominator of that model; the other four models have valid runs on all 105 tasks.

\begin{figure}
  \centering
  \includegraphics[width=\textwidth]{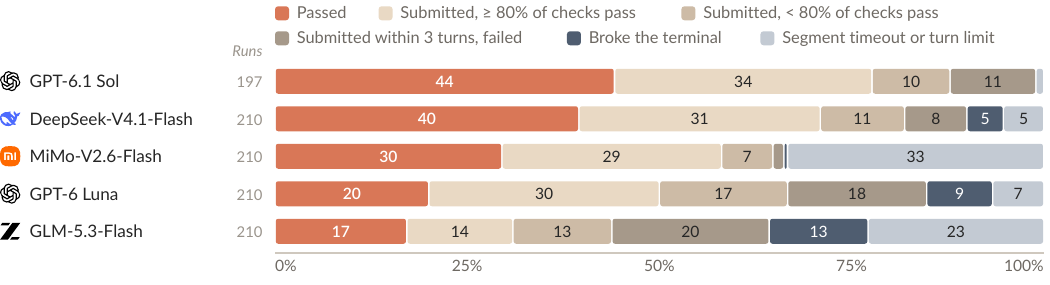}
  \caption{Outcome of every run without a cap on our tasks, as a share of each model's runs. Failed runs are classed by how they stop.}
  \label{fig:fails}
\end{figure}

\textbf{Failure modes.} \autoref{fig:fails} classifies every run without a cap by how it ends. GPT-6 Luna and GLM-5.3-Flash often submit within three turns without verifying their work (18\% and 20\%), a failure that the taxonomy of agent failures in Terminal-Bench also lists~\citep{merrill2026terminalbench}. GLM-5.3-Flash also breaks the terminal session in 13\% of its runs. MiMo-V2.6-Flash rarely submits early but spends long on each turn and reaches the time limit of a segment or the turn limit in 33\% of its runs.

The three cases below show how runs end early or run out of time.

\begin{casebox}{Submitting without checking (Pacemaker, segment 1, GPT-6 Luna)}
Pacemaker manages resources in Linux high-availability clusters. The instruction asks for five fault-injection operations in a given order, with their output saved to a log. GPT-6 Luna sent all five commands in one turn. The first lacked the flag for automatic confirmation and stopped to wait for input, which swallowed the commands after it, so the log recorded the operations in the wrong order. GPT-6 Luna ended the segment without reading the output.
\end{casebox}

\begin{casebox}{Breaking the terminal (Laravel application, segment 3, GLM-5.3-Flash)}
GLM-5.3-Flash had a complete plan but began a batch of commands with \texttt{set -e}, which makes the shell exit as soon as a command fails. In an interactive shell the setting persists: a later command returned a nonzero status, the shell exited, and the session ended together with the remaining segments. Most broken sessions of GLM-5.3-Flash and GPT-6 Luna come from \texttt{set -e} or \texttt{exit}.
\end{casebox}

\begin{casebox}{Reaching the time limit (Spark with Spooq, segment 1, MiMo-V2.6-Flash)}
The instruction asks to run a sample job in the Spark shell and save its output. Two problems stand in the way: the wrong entry point causes a null-pointer exception, and the job lowers the log level to WARN, which hides the line the output needs. MiMo-V2.6-Flash found the first cause, then kept reading bytecode to trace the second without first saving an output that would pass, and ran out of time.
\end{casebox}

\section{Training Data Details}
\label{app:train}

At evaluation, terminus-2 compresses the context through three calls: the model summarizes the work so far, asks questions in a new context that holds the instruction, the summary and the current screen, and answers them from the old context. The runs we collect come from long-context models and contain no compression, so a model trained on them would never see what follows one; we therefore cut each run where terminus-2 would compress it. The sample after a cut starts with the messages terminus-2 would give the model, with the summary, questions and answers generated by an LLM and the instruction covering all segments delivered so far. The three calls also become samples with prose targets. Replies the harness cannot parse are kept with its error message but excluded from the loss.

\end{document}